\documentclass[11pt,a4paper]{article}
\usepackage{times,latexsym}
\usepackage{url}
\usepackage[T1]{fontenc}
\usepackage[sort,round]{natbib}
\usepackage{fullpage}
\usepackage[hidelinks]{hyperref}

\usepackage{tikz}

\usepackage{amsmath}
\usepackage{amsfonts}
\usepackage{amssymb}
\usepackage{xcolor}
\usepackage{textcomp}
\usepackage{booktabs}
\usepackage{multirow}
\usepackage{colortbl}
\usepackage{enumitem}
\usepackage{subcaption}
\usepackage{xspace}
\usepackage{pdfpages}
\usepackage{soul}

\newcommand{\specialcell}[2][c]{\footnotesize\begin{tabular}[#1]{@{}l@{}}#2\end{tabular}}
\usepackage{amsmath,amsfonts,bm}

\def\eqref#1{equation~\ref{#1}}

\def\1{\bm{1}}

\DeclareMathAlphabet{\mathsfit}{\encodingdefault}{\sfdefault}{m}{sl}
\SetMathAlphabet{\mathsfit}{bold}{\encodingdefault}{\sfdefault}{bx}{n}

\newcommand{\R}{\mathbb{R}}

\newcommand{\bx}{\mathbf{x}}
\newcommand{\by}{\mathbf{y}}
\newcommand{\btheta}{\boldsymbol{\theta}}
\newcommand{\bphi}{\boldsymbol{\phi}}

\newcommand{\test}{\mathrm{tst}}
\newcommand{\train}{\mathrm{tr}}
\newcommand{\pretrain}{\mathrm{ptr}}
\newcommand{\basestr}{\mathrm{base}}

\newcommand{\ytest}{\by_{\test}}
\newcommand{\xtest}{\bx_{\test}}
\newcommand{\xtrain}{\bx_{\train}}
\newcommand{\ytrain}{\by_{\train}}
\newcommand{\xpretrain}{\bx_{\pretrain}}
\newcommand{\ypretrain}{\by_{\pretrain}}
\newcommand{\xbase}{\bx_{\basestr}}
\newcommand{\ybase}{\by_{\basestr}}
\newcommand{\btau}{\boldsymbol{\tau}}

\newcommand{\datasetx}{\mathcal{X}}
\newcommand{\datasety}{\mathcal{Y}}
\newcommand{\datasetytest}{\datasety_{\test}}
\newcommand{\datasetxtest}{\datasetx_{\test}}
\newcommand{\datasetxtrain}{\datasetx_{\train}}
\newcommand{\datasetytrain}{\datasety_{\train}}
\newcommand{\datasetxpretrain}{\datasetx_{\pretrain}}
\newcommand{\datasetypretrain}{\datasety_{\pretrain}}

\newcommand{\thetaopt}{\btheta^{*}}
\newcommand{\thetainitial}{\hat{\btheta}}

\usepackage[noabbrev,capitalize,nameinlink]{cleveref}
\crefformat{section}{\S#2#1#3}  
\crefname{section}{\S}{\S\S}
\Crefname{section}{\S}{\S\S}
\crefname{table}{Table}{}
\crefname{figure}{Figure}{}
\crefname{algorithm}{Algorithm}{}
\crefname{equation}{Equation}{}
\crefname{appendix}{Appendix}{}
\crefname{thm}{Theorem}{}
\crefname{prop}{Proposition}{}
\crefname{cor}{Corollary}{}
\crefname{observation}{Observation}{}
\crefname{assumption}{Assumption}{}
\crefformat{section}{\S#2#1#3}

\newcommand{\Nentries}{619\xspace}
\newcommand{\Npapers}{449\xspace}
\newcommand{\overNpapers}{400\xspace}
\newcommand{\overNstudies}{600\xspace}

\title{State-of-the-art generalisation research in NLP:\\{\Large A taxonomy and review}}

\newcommand{\fb}[0]{$^{1}$}
\newcommand{\amsterdam}[0]{$^{2}$}
\newcommand{\zurich}[0]{$^{3}$}
\newcommand{\edinburgh}[0]{$^{4}$}
\newcommand{\reka}[0]{$^{5}$}
\newcommand{\allenAI}[0]{$^{6}$}
\newcommand{\washington}[0]{$^{7}$}
\newcommand{\cambridge}[0]{$^{8}$}
\newcommand{\amazon}[0]{$^{9}$}
\newcommand{\ens}[0]{$^{10}$}
\newcommand{\wbank}[0]{$^{11}$}
\newcommand{\harvard}[0]{$^{12}$}
\newcommand{\aberdeen}[0]{$^{13}$}
\newcommand{\copenhagen}[0]{$^{14}$}
\newcommand{\pioneer}[0]{$^{15}$}
\newcommand{\tshuttle}[0]{$^{16}$}
\newcommand{\huygaa}[0]{$^{17}$}
\newcommand{\hongkong}[0]{$^{18}$}
\newcommand{\MIT}[0]{$^{19}$}
\newcommand{\mpi}[0]{$^{20}$}

\author{
	Dieuwke Hupkes\fb, Mario Giulianelli\amsterdam$^{,}$\zurich, Verna Dankers\fb$^{,}$\edinburgh, Mikel Artetxe\reka\\
	Yanai Elazar\allenAI$^{,}$\washington, Tiago Pimentel\cambridge, Christos Christodoulopoulos\amazon,  Karim Lasri\ens$^{,}$\wbank\\
	Naomi Saphra\harvard,
	Arabella Sinclair\aberdeen, Dennis Ulmer\copenhagen, Florian Schottmann\zurich$^{,}$\tshuttle\\
	Khuyagbaatar Batsuren\huygaa, Kaiser Sun\fb, Koustuv Sinha\fb, Leila Khalatbari\hongkong\\
	Maria Ryskina\MIT, Rita Frieske\hongkong, Ryan Cotterell\zurich, Zhijing Jin\zurich$^{,}$\mpi\\
	\\
	\texttt{dieuwkehupkes@meta.com} \hspace{6mm} \texttt{mgiulianelli@inf.ethz.ch}\\
	\texttt{vernadankers@gmail.com}\\
	\\
	{\footnotesize
		\fb FAIR \hspace{1mm}
		\amsterdam University of Amsterdam \hspace{1mm}
		\zurich ETH Z\"{u}rich \hspace{1mm}
		\edinburgh University of Edinburgh
	}\\ 
	{\footnotesize
		\reka Reka AI \hspace{1mm}
		\allenAI Allen Institute for AI \hspace{1mm}
		\washington University of Washington \hspace{1mm}
		\cambridge University of Cambridge
	}\\
	{\footnotesize
		\amazon Amazon Alexa AI  \hspace{1mm}
		\ens \'{E}cole Normale Sup\'{e}rieure-PSL \hspace{1mm}
		\wbank The World Bank \hspace{1mm}
		\harvard Harvard University
	}\\
	{\footnotesize
		\aberdeen University of Aberdeen \hspace{1mm}
		\copenhagen IT University of Copenhagen \hspace{1mm}
		\pioneer Pioneer Centre for Artificial Intelligence
	}\\
	{\footnotesize
		\tshuttle Textshuttle \hspace{1mm}
		\huygaa National University of Mongolia \hspace{1mm}
		\hongkong Hong Kong University of Science and Technology
	}\\
	{\footnotesize
		\MIT MIT \hspace{1mm}
		\mpi Max Planck Institute for Intelligent Systems
	}
}

\date{}

\begin{document}
\maketitle

\begin{abstract}
    The ability to generalise well is one of the primary desiderata of natural language processing (NLP).
    Yet, what `good generalisation' entails and how it should be evaluated is not well understood, nor are there any evaluation standards for generalisation.
    In this paper, we lay the groundwork to address both of these issues.
    We present a taxonomy for characterising and understanding generalisation research in NLP.
    Our taxonomy is based on an extensive literature review of generalisation research, and contains five axes along which studies can differ: their main motivation, the type of generalisation they investigate, the type of data shift they consider, the source of this data shift, and the locus of the shift within the modelling pipeline.
    We use our taxonomy to classify over \overNpapers papers that test generalisation, for a total of more than \overNstudies individual experiments.
    Considering the results of this review, we present an in-depth analysis that maps out the current state of generalisation research in NLP, and we make recommendations for which areas might deserve attention in the future.
    Along with this paper, we release a webpage where the results of our review can be dynamically explored, and which we intend to update as new NLP generalisation studies are published.
    With this work, we aim to take steps towards making state-of-the-art generalisation testing the new status quo in NLP.\looseness-1
    \vspace{1em}

    \centering
   	\fbox{
   		\begin{minipage}{11.7cm}
   			\begin{center}
	  			This preprint was published as an \href{https://doi.org/10.1038/s42256-023-00729-y}{{\color{blue}\ul{Analysis article}}} in Nature Machine Intelligence. Please refer to the published version when \href{https://www.nature.com/articles/s42256-023-00729-y\#citeas}{{\color{blue}\ul{citing this work}}}.
  			\end{center}
   		\end{minipage}
   	}

\end{abstract}

\section{Introduction}

Good generalisation, roughly defined as the ability to successfully transfer representations, knowledge, and strategies from past experience to new experiences,
is one of the primary desiderata for models of natural language processing (NLP), as
well as for models in the wider field of machine learning \citep[i.a.]{marcus1998rethinking,schmidhuber1990towards,wong2007generalisation,lake2017building,yogatama2019learning,linzen-2020-accelerate,elangovan-etal-2021-memorization, marcus2018deep,kirk-etal-2021-survey,shen2021towards}.
For some, generalisation is crucial to ensure that models behave robustly, reliably, and fairly when making predictions about data different from the data that they learned from, which is of critical importance when models are employed in the real world.
Others see good generalisation as intrinsically equivalent to good performance and believe that without it a model is not truly able to conduct the task we intend it to.
Yet others strive for good generalisation because they believe models should behave in a human-like way, and humans are known to generalise well.
While the importance of generalisation is almost undisputed
-- in the past five years, in the ACL Anthology alone over 1200 papers mentioned it in their title or abstract -- systematic generalisation testing is not the status quo in the field of NLP. 

At the root of this problem lies the fact that there is little understanding and agreement about what good generalisation looks like, what types of generalisation exist, and which should be prioritised in varying scenarios.
Broadly speaking, generalisation is evaluated by assessing how well a model performs on a test dataset, given the relationship of this dataset with the data the model was trained on.
For decades, it was common to exert only one simple constraint on this relationship: that the train and test data are different.
Typically, this was achieved by randomly splitting available data into a training and a test partition.
Generalisation was thus evaluated by training and testing models on different but similarly sampled data, assumed to be independent and identically distributed (i.i.d.).
In the past 20 years, we have seen great strides on such random train--test splits in a range of different applications.
Since the first release of the Penn Treebank \citep{marcus-etal-1993-building}, $F_1$ scores for labelled constituency parsing went from above 80\% at the end of the previous century \citep{magerman-1995-statistical,collins-1996-new} and close to 90\% in the first ten years of the current one \citep[e.g.][]{petrov-klein-2007-improved,sangati-zuidema-2011-accurate} to scores up to 96\% in recent years \citep{mrini-etal-2020-rethinking,yang2020strongly}.
On the same corpus, performance for language modelling went from per-word perplexity scores
well above 100 in the mid-90s \citep{rosenfeld1996maximum,kneser1995improved} to a score of 20.5 in 2020 \citep{brown2020language}.
In many areas of NLP, the rate of progress has become even faster in the recent past.  
Scores for the popular evaluation suite GLUE went from values between 60 and 70 at its release in 2018 \citep{wang-etal-2018-glue} to scores exceeding 90 less than a year after \citep{devlin-etal-2019-bert}, with performances on a wide range of tasks reaching and surpassing human-level scores by 2019 \citep[e.g.][]{wang-etal-2018-glue,wang2019super,devlin-etal-2019-bert,liu2019roberta}.
In~2022, strongly scaled-up models \citep[e.g.][]{chowdhery2022palm} showed astounding performances on almost all existing i.i.d.\ natural language understanding benchmarks.

With this progress, however, came the realisation that, for an NLP model, reaching very high or human-level
scores on an i.i.d.\ test set does not imply that the model robustly generalises to a wide range of different scenarios in the way humans do.
In the recent past, we witnessed a tide of different studies pointing out generalisation failures in neural models that have state-of-the-art scores on random train--test splits \citep[to give just a few examples]{blodgett-etal-2016-demographic,plank2016non,marcus2018deep,lake2018generalization,mccoy-etal-2019-right,kim-linzen-2020-cogs,sinha-etal-2021-masked,khishigsuren2022using,razeghi2022impact}.
Some show that when models perform well on i.i.d.\ test splits, they might rely on simple heuristics that do not robustly generalise in a wide range of non-i.i.d.\ scenarios \citep{mccoy-etal-2019-right,kaushik2019learning,gardner-etal-2020-evaluating}, over-rely on stereotypes \citep{parrish-etal-2022-bbq,srivastava2022beyond}, or bank on memorisation rather than generalisation \citep{razeghi2022impact,lewis-etal-2021-question}.
Others, instead, display cases in which performances drop when the evaluation data differs from the training data in terms of genre, domain or topic \citep[e.g.][]{plank2016non,michel-neubig-2018-mtnt,malinin2021shifts}, or when it represents different subpopulations \citep[e.g.][]{blodgett-etal-2016-demographic,dixon2018measuring}.
Yet other studies focus on models' inability to generalise compositionally \citep{lake2018generalization,kim-linzen-2020-cogs,li-etal-2021-compositional,dankers-etal-2022-paradox}, structurally \citep{sinha-etal-2021-masked,wei-etal-2021-frequency,weber-etal-2021-language}, to longer sequences \citep{raunak2019compositionality,dubois-etal-2020-location}, or to slightly different task formulations of the same problem  \citep{srivastava2022beyond}.

By showing that good performance on traditional train--test splits does not equal good generalisation, the examples above bring into question what kind of model capabilities recent breakthroughs actually reflect, and they suggest that research on the evaluation of NLP models is catching up with the fast recent advances in architectures and training regimes.
Unfortunately, this body of work also reveals that there is no real agreement on what kind of generalisation is important for NLP models: different studies encompass a wide range of generalisation-related research questions, and they use a wide range of different methodologies and experimental setups.
As of yet, it is unclear how the results of different studies relate to each other: how should generalisation be assessed, if not with i.i.d.\ splits?
How do we determine what types of generalisation are already well addressed and which are neglected, or which types of generalisation should be prioritised?
Ultimately, on a meta-level, how can we provide answers to these important questions without a systematic way to discuss generalisation in NLP?
These missing answers are standing in the way of better model evaluation and model development: what we cannot measure, we cannot improve.

\begin{figure}[]
    \centering
    \includegraphics[width=\textwidth]{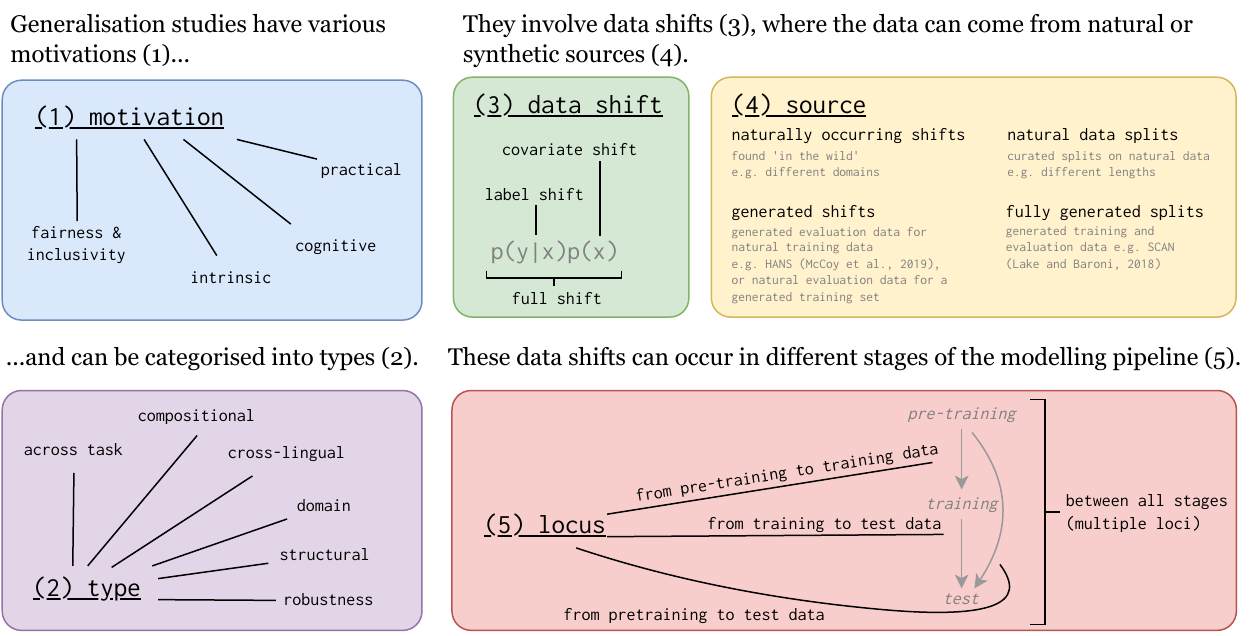}
    \caption{A graphical representation of our proposed taxonomy of generalisation in NLP. The taxonomy consists of five different (nominal) axes that describe the high-level \emph{motivation} of the work (\cref{sec:motivations}), the \emph{type} of generalisation the test is addressing (\cref{sec:generalisation_target}), what kind of \emph{data shift} occurs between training and testing (\cref{sec:data_shifts}), and what the \emph{source} and \emph{locus} of this shift are (\cref{sec:shift_sources} and \cref{sec:shift_locus}, respectively).
    }\label{fig:taxonomy-infographic}

\end{figure}

The current article introduces a new framework to systematise and understand generalisation research, and it is an attempt to provide answers to the questions above.
We present a \textbf{generalisation taxonomy}, a \textbf{meta-analysis} of existing research on generalisation in NLP, a set of \textbf{online tools} that can be used by researchers to explore and better understand generalisation studies through \href{https://genbench.github.io}{\underline{\textcolor{blue}{our website}}}, and we introduce \textbf{evaluation cards} that authors can use to comprehensively summarise the generalisation experiments conducted in their papers.
We believe that \textit{state-of-the-art generalisation testing} should be the new status quo in NLP, and with this work, we aim to lay the groundwork for facilitating this change.
In the remainder of this article, we first describe the five axes of our taxonomy (\cref{sec:motivations}-\ref{sec:shift_locus}); these are the main axes along which generalisation studies differ.
In \cref{sec:survey}, we present our analysis of the current state of generalisation research, grounded on a review of \Npapers papers and a total of \Nentries generalisation experiments.
In \cref{sec:conclusion}, we summarise our main findings and make concrete recommendations for more sound and exhaustive generalisation tests in NLP research.

\section{The generalisation taxonomy}

We now begin a discussion of the five axes of the proposed generalisation taxonomy, which are also visualised in \cref{fig:taxonomy-infographic} and summarised in \cref{appendix:taxonomy_summary}.
The proposed taxonomy intends to be beneficial to understanding generalisation research in NLP \textit{in hindsight} but is also meant as an active device for characterising ongoing studies as well as work that is still to come.
We facilitate this through \emph{evaluation cards} -- analogous to the model cards proposed by \citet{mitchell2019model} and the data sheets of \citet{gebru2021datasheets} -- which researchers can fill out for the experiments they conducted in their work and include in their paper.
Doing so aids the cause of making generalisation evaluation the status quo, and enables effective monitoring of trends in generalisation research.
An example of an evaluation card is provided in \cref{fig:eval_card}; \cref{sec:eval_cards} elaborates on how to use the cards.

\begin{figure}[]
    \centering
\newcommand{\tabularwidth}{\textwidth}

\newcommand{\firstexp}{$\mathbf{\square}$}
\newcommand{\secondexp}{$\mathbf{\bigtriangleup}$}
\newcommand{\thirdexp}{$\mathbf{\bigcirc}$}

\renewcommand{\arraystretch}{1.1}
\setlength{\tabcolsep}{0mm}
\begin{tabular}{|p{\tabularwidth}<{\centering}|}
    \hline

    \rowcolor{gray!60}
    \textbf{Motivation} \\
    \begin{tabular}{
        p{0.25\tabularwidth}<{\centering}
        p{0.25\tabularwidth}<{\centering}
        p{0.25\tabularwidth}<{\centering}
        p{0.25\tabularwidth}<{\centering}}
    \textit{Practical} & \textit{Cognitive} & \textit{Intrinsic} & \textit{Fairness}\\
    \large
        \firstexp \hspace{2mm} \secondexp \hspace{2mm} \thirdexp      
    &                                   
    &                                   
    &                                   
    \vspace{2mm} \\ 
    \end{tabular}\\

    \rowcolor{gray!60}
    \textbf{Generalisation type} \\
    \begin{tabular}{
        p{0.21\tabularwidth}<{\centering}
        p{0.2\tabularwidth}<{\centering}
        p{0.13\tabularwidth}<{\centering}
        p{0.13\tabularwidth}<{\centering}
        p{0.13\tabularwidth}<{\centering}
        p{0.2\tabularwidth}<{\centering}}
    \textit{Compositional} & \textit{\hspace{4mm}Structural} & \textit{Task} & \textit{Language} & \textit{Domain} & \textit{Robustness}\\
    \large
        &                               
        & \large\firstexp                     
        & \large\secondexp                    
        & \large\thirdexp                     
        &                               
        \vspace{2mm}\\
    \end{tabular}\\

    \rowcolor{gray!60}
    \textbf{Shift type} \\
    \begin{tabular}{
        p{0.25\tabularwidth}<{\centering}
        p{0.25\tabularwidth}<{\centering}
        p{0.25\tabularwidth}<{\centering}
        p{0.25\tabularwidth}<{\centering}}
        \textit{Covariate} & \textit{Label} & \textit{Full} & \textit{Assumed}\\
        \large
         \thirdexp \secondexp           
         & \large \secondexp\hspace{2mm} \firstexp                    
         &                              
         &                              
         \vspace{2mm}\\
    \end{tabular}\\

    \rowcolor{gray!60}
    \textbf{Shift source} \\
    \begin{tabular}{
        p{0.28\tabularwidth}<{\centering}
        p{0.28\tabularwidth}<{\centering}
        p{0.22\tabularwidth}<{\centering}
        p{0.22\tabularwidth}<{\centering}}
        \textit{Naturally occuring} & \textit{Partitioned natural} & \textit{Generated shift} & \textit{Fully generated}\\
        \large
        \firstexp\hspace{2mm}  \secondexp \hspace{2mm} \thirdexp  
        &                               
        &                               
        &                               
        \vspace{2mm}\\
    \end{tabular}\\

    \rowcolor{gray!60}
    \textbf{Shift locus}\\
    \begin{tabular}{
        p{0.25\tabularwidth}<{\centering}
        p{0.25\tabularwidth}<{\centering}
        p{0.25\tabularwidth}<{\centering}
        p{0.25\tabularwidth}<{\centering}}
        \textit{Train--test} & \textit{Finetune train--test} & \textit{Pretrain--train} & \textit{Pretrain--test}\\
        & \large \secondexp \hspace{2mm} \thirdexp          
        & \large \secondexp                    
        & \large \firstexp                     
        \vspace{2mm}\\
    \end{tabular}\\
    \hline
\end{tabular}

    \caption{Example of evaluation card that can be used to summarise all experiments in a paper. Authors can mark where on the five taxonomy axes their experiments belong, as is illustrated with symbols for three hypothetical experiments in this figure. In \cref{sec:eval_cards}, we will further discuss how to use the evaluation cards, and we provide also a single-column version of it. On our website, we provide a tool to automatically generate latex code for evaluation cards.
}\label{fig:eval_card}
\end{figure}

\subsection{Motivation: what is the high-level motivation for a generalisation test?}
\label{sec:motivations}

The first axis we consider is the high-level motivation of a generalisation study. 
We identified four closely intertwined goals of generalisation research in NLP, which we refer to as the \textit{practical}, the \textit{cognitive}, the \textit{intrinsic}, and the \textit{fairness} motivation.
The motivation of a study determines what type of generalisation is desirable, it shapes the experimental design, and it affects which conclusions can be drawn from a model's display or lack of generalisation.
It is therefore crucial for researchers to be explicitly aware of the motivation underlying their studies to ensure that the experimental setup aligns with the questions they seek to answer.%
\footnote{
As we will see in what follows, the same questions can often be asked with different underlying motivations.
This makes it sometimes difficult to identify what exactly the motivation of a generalisation study is.
Often, studies may inform conclusions along all four dimensions.
However, given the importance of the motivation for the implications and design of the study, we nevertheless try to identify the main guiding motive of a study in our review (\cref{sec:survey}), and we encourage researchers to be explicit about the motivation of their future studies.}

\subsubsection{Practical: in what settings can the model be used or improved?}
One frequent motivation to study generalisation is of a markedly practical nature.
Studies that consider generalisation from a practical perspective seek to assess in what kind of scenarios a model can be deployed, or which modelling changes can improve performance in various evaluation scenarios.
An example of a research question that is often addressed with a primarily practical motivation is how well models generalise to different text domains or to data collected in different ways.
For instance, \citet{michel-neubig-2018-mtnt} consider how well machine translation models trained on canonical text can generalise to noisy data from an internet platform, and \citet{lazaridou2021mind} investigate language model generalisation to texts written in different time periods.
Other questions that are frequently addressed from a practical perspective concern biases in the training data, and whether models robustly generalise to datasets that do not share those biases, or whether they learnt spurious correlations due to that bias \citep[e.g.][]{zhou-etal-2021-hidden,behnke-etal-2022-bias}.

\subsubsection{Cognitive: does the model generalise like a human?}
A second high-level motivation that drives generalisation research is cognitively oriented and can be separated into two underlying categories: one focusing on models and one aimed at learning about cognition and the language faculty in humans through computational models.
The first category is related to model behaviour: human generalisation is a useful reference point for the evaluation of models in NLP because it is considered to be a hallmark of human intelligence
\citep[e.g.][]{lake2017building,marcus2003algebraic} and, perhaps more importantly, because it is precisely the type of generalisation that is required to successfully model natural language.
Humans learn quickly, from fewer data than existing models, and they easily (compositionally) recombine concepts they already know to understand concepts they have never before encountered \citep{fodor1988connectionism,linzen-2020-accelerate,marcus2018deep}.
These feats are thus, arguably, important desiderata for models.\footnote{
	We do not always expect from a model the same type or level of generalisation a human exhibits.
  There are cases in which it is desirable for models to generalise better than humans, for example across languages -- something humans typically do not excel at.
  In other cases, such as language identification, models already generalise better than humans and would hardly be useful if they did not.}
In some cases, it might be difficult to distinguish cognitive from practical motivations:
a model that generalises like a human should score well also on practically motivated tests, which is why the same experiments can be motivated in multiple ways.
In our axes-based taxonomy,
we rely on the motivations provided by the authors.
Compositional generalisation experiments, for instance, can be cognitively motivated -- e.g.\ when the authors suggest machines ought to generalise the way humans do -- but also practically -- e.g.\ when the authors question which machine learning techniques improve performance on benchmarks that happen to be used to test compositional generalisation.\looseness-1

The second, more deeply cognitively inspired category embraces work that evaluates generalisation in models to learn more about language and cognition \citep[e.g.][]{marcus1999connectionism,hupkes2020hierarchy,baroni2021proper,mcclelland1999does,lakretz2021mechanisms}.
Studies in this category investigate what underlies generalisation in computational models, not in order to improve the models' generalisation capabilities but to derive new hypotheses about the workings of human generalisation.

\subsubsection{Intrinsic: does the model solve the task correctly?}
A third motivation to evaluate generalisation in NLP models, which cuts through the two previous motivations, appertains to the question of whether models learned the task we intended them to learn, in the way we intended the task to be learned.
The shared presupposition underpinning this type of research is that if a model has truly learned the task it is trained to do, it should be able to execute this task also in settings that differ from the exact training scenarios.
What changes, across studies, is the set of conditions under which a model is considered to have appropriately learned a task.
Researchers studying compositional generalisation (see \cref{subsec:compositional}), for example, assume that a correct understanding of language implies that the underlying compositional structure of language is captured; under that assumption, a model should not have trouble generalising to new inputs that are generated using the same compositional system.
Others instead argue that true language understanding implies being able to use language across a wide variety of tasks (see \cref{subsec:crosstask}).
Yet others maintain that for a model to truly capture aspects of language understanding, such as relations of entailment between two sentences \citep[e.g.][]{marelli-etal-2014-sick,bowman-etal-2015-large,williams-etal-2018-broad}, it should be able to do so across different datasets, even if those were sampled in a slightly different way \citep[e.g.][]{talman-chatzikyriakidis-2019-testing}.
In studies that consider generalisation from this perspective, generalisation failures are taken as proof that the model -- in fact -- did not learn the task as we intended it to learn it. Instead, it displayed behaviour that superficially made us think it did, for instance by relying on spurious patterns or non-generalisable heuristics.\looseness-1

\subsubsection{Fairness and inclusivity: does the model generalise in a fair and responsible way?}
A last yet very important motivation for generalisation research is the desire to have models that are fair, responsible and unbiased.
One category of studies driven by these concepts, often ethical in nature, asks questions about how well models generalise to diverse demographics, typically considering minority or marginalised groups \citep[e.g.][]{blodgett-etal-2016-demographic,bender2021stochastic,koh2021wilds}, or investigates to what extent models perpetuate (undesirable) biases learned from their training data \citep[e.g.][]{dixon2018measuring,hutchinson-etal-2020-social,sheng-etal-2019-woman}.
Another line of research related to both fairness and inclusivity focuses on efficiency, both in terms of the amount of data that is required for a model to converge to a solution as well as the necessary amount of compute.
In such studies, efficiency is seen \textit{as a correlate} of generalisation: models that generalise well should learn more quickly and require fewer data \citep[see, e.g.][]{marcus2018deep}. As such, they
are more inclusively applicable -- for instance to low-resource languages or demographic groups for which little data is available,
they are more accessible for groups with smaller computational resources, and they have a lower environmental impact \citep[see, e.g.][]{strubell-etal-2019-energy}.
While they have not been mentioned before in this section, studies on efficiency can naturally also be motivated by practical concerns, as well as by cognitive interests (e.g.\ comparing sample efficiency in humans and models).\looseness-1

\subsection{Generalisation type: what type of generalisation is a test addressing?}
\label{sec:generalisation_target}

\begin{figure}
    \centering
\includegraphics[width=1.0\textwidth]{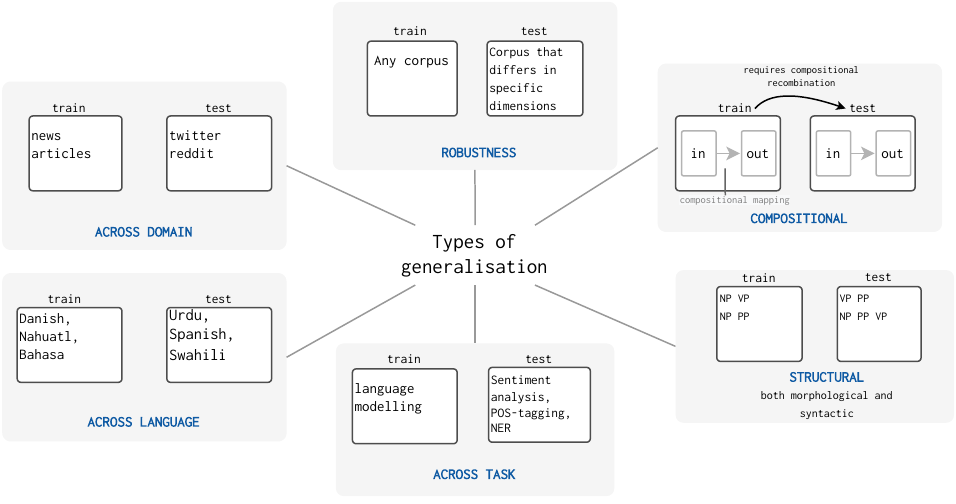}
    \caption{The six generalisation types, explained in detail in \cref{subsec:compositional}-\cref{subsec:robustness}.}\label{fig:type_infographic}
\end{figure}

The second axis in our taxonomy describes, on a high level, what aspects of generalisation a test is intended to capture,
making it an important axis of our taxonomy.
We identify and describe six types of generalisation that are frequently considered in the literature.
Some types are rooted in knowledge about human generalisation, such as those that target \textit{compositional} (\cref{subsec:compositional}) or \textit{structural} generalisation (\cref{subsec:structural}).
Others, instead, are motivated by more practical concerns, such as generalisation across \textit{tasks} (\cref{subsec:crosstask}), \textit{languages} (\cref{subsec:crosslingual}) and \textit{domains} (\cref{subsec:domain}), or by an interest in analysing how \textit{robustly} models generalise (\cref{subsec:robustness}).
An overview of generalisation types is presented in Figure~\ref{fig:type_infographic}.\looseness-1

\subsubsection{Compositional generalisation}
\label{subsec:compositional}

The first prominent type of generalisation addressed in the literature is \emph{compositional generalisation}, which is often argued to underpin human's ability to quickly generalise to new data, tasks and domains \citep{fodor1988connectionism,lake2017building,schmidhuber1990towards,marcus2018deep}.
Because of this strong connection with humans and human language, work on compositional generalisation often has a primarily cognitive motivation, although practical concerns such as sample efficiency, quick adaptation and good generalisation in low-resource scenarios are frequently mentioned as additional or alternative drivers \citep[to give just a few examples]{linzen-2020-accelerate,chaabouni-etal-2021-transformers}.
While it has a strong intuitive appeal and clear mathematical definition \citep{montague1970universal}, compositional generalisation is not easy to pin down empirically.
Here, we follow \citet{schmidhuber1990towards} in defining compositionality as the ability to systematically recombine previously learned elements to map new inputs made up from these elements to their correct output.\footnote{
For an elaborate account of the different arguments that come into play when defining and evaluating compositionality for a neural network, we refer to \citet{hupkes2020compositionality}.
}
In language, the inputs are `forms' (e.g.\ phrases, sentences, larger pieces of discourse), which are mapped to their meaning or interpretation.
Since compositional generalisation is defined in terms of both an input and output space, it is usually evaluated in tasks such as sequence classification \citep[e.g.][]{bowman2015tree,hupkes2018diagnostic,veldhoen2016diagnostic}, machine translation \citep[e.g.][]{raunak2019compositionality,dankers-etal-2022-paradox,liu2021challenges}, semantic parsing \citep[e.g.][]{shaw-etal-2021-compositional,finegan-dollak-etal-2018-improving,kim-linzen-2020-cogs,keysers2019measuring} or other kinds of generative tasks \citep[e.g.][]{lake2018generalization,hupkes2020compositionality}.
In such tasks, the input and output spaces are clearly distinct. 
As far as we are aware, there have not yet been many explicit systematic attempts to evaluate compositionality in (ungrounded) language models.\footnote{
	There are, however, several studies that focus on \emph{structural} generalisation in such models. Contrary to compositional generalisation, structural generalisation does not focus on the ability of models to correctly interpret new inputs, or to assign meanings to them, but only on their ability to generalise with respect to input forms. We discuss structural generalisation in \cref{subsec:structural}.\looseness-1}
If and how compositionality can be adequately evaluated in such models, where the input and output (form and meaning) are conflated in one space (the space defined by the language vocabulary), are questions that are yet to be answered.\footnote{
	An interesting example of this open research line is the qualitative study conducted by \citet{brown2020language} to test if GPT-3 can use novel words correctly in a sentence; as another example, slightly further away from traditional forms of compositionality, \citet{talmor-etal-2020-olmpics} finetune pretrained masked language models on multi-hop composition in question answering.}\looseness-1

\subsubsection{Structural generalisation}
\label{subsec:structural}

A second category of usually cognitively inspired generalisation studies focuses on the extent to which models can process or generate structurally (grammatically) correct forms, rather than on whether they can assign to forms correct interpretations.
Unlike compositional generalisation, structural generalisation does not require an output space (the meaning or interpretation space; see \cref{subsec:compositional}). This makes it more straightforwardly evaluated in form-only models (i.e.\ language models).
We distinguish two broad categories of structural generalisation: syntactic generalisation and morphological generalisation.

\paragraph{Syntactic generalisation} Some structural generalisation studies focus specifically on \textit{syntactic generalisation}: they consider whether models can generalise to novel syntactic structures or novel elements in known syntactic structures.
The typical experimental setup involves training data designed to contain or exclude specific conditions: \citet{jumelet-etal-2021-language} and \citet{weber-etal-2021-language} remove specific grammatical environments from the training data
and then test whether models nevertheless learn to generalise to such environments;
\citet{wei-etal-2021-frequency} vary word frequencies in the training corpus to investigate how syntactic rule learning in pretrained language models is affected by the frequencies observed in the training phase. 
It is unfortunately difficult to conduct this type of study using very large language models: the computational cost of training these models on multiple datasets is high and generating specific test splits given knowledge of what is in the training data is often not possible, as large models' training data is often not in the public domain.
Overall, the lack of control over the relationship between the training and the evaluation data of large language models makes it hard to assess to what extent the incidental examples reported for these models (most notably, in their respective release papers) are reflective of successful generalisation. This problem has only very recently begun to be acknowledged in the NLP community, with models being now openly released together with their training data \citep[e.g.][]{zhang2022opt,scao2022bloom}.

\paragraph{Morphological generalisation} A second category of structural generalisation studies focuses on morphological inflection, a popular testing ground for questions about human structural generalisation abilities.
Papers focusing on morphological inflection \citep[e.g.][]{liu-hulden-2022-transformer,mccurdy-etal-2020-inflecting,malouf2017abstractive,dankers-etal-2021-generalising,kirov-cotterell-2018-recurrent,corkery-etal-2019-yet} are typically rooted in strong cognitive motivations.
While most of this work considers i.i.d.\ train--test splits, recent studies have focused on how morphological transducer models generalise across languages \citep[e.g.][]{mccarthy-etal-2019-sigmorphon,vylomova-etal-2020-sigmorphon,pimentel-ryskina-etal-2021-sigmorphon}
as well as within each language \citep{liu-hulden-2022-transformer,pimentel-etal-2021-sigmorphon,wilson-etal-2021-were,szolnok-barta-lakatos-etal-2021-bme,calderone-etal-2021-not-quite,li-etal-2021-leveraging}.
These studies often take inspiration from the so-called \emph{wug} tests used in psycholinguistics to assess human morphological generalisation to novel words \citep{berko1958child,marcus1995german}.
They can potentially also be conducted with large language models but the lack of access to their training data, as explained before, makes it difficult to determine whether the supposedly novel test words were truly never seen by the models.

\subsubsection{Generalisation across tasks}
\label{subsec:crosstask}

A third direction of generalisation research considers the ability of individual models to adapt to multiple NLP problems.
We refer to this ability as generalisation across tasks or cross-task generalisation.
Along with the great advancements in NLP models, in the past ten years, the nature of cross-task generalisation tests has changed quite substantially; we discuss this evolution in the current section.

\paragraph{Multitask learning}
Cross-task generalisation in NLP has been traditionally strongly connected to transfer and multitask learning \citep{collobert2008unified}.
In multitask learning, a model is either trained and evaluated on a set of tasks, or pretrained on some tasks and then adapted to others.
As this setup favours approaches that benefit from positive transfer across tasks, it implicitly studies forms of cross-task generalisation.\footnote{
	As illustrated by the work of \citet{weber-etal-2021-language}, the definition of \emph{task} can be taken liberally in this context, ranging from traditional notions of NLP tasks to considering subproblems of a single classic NLP task. For instance, while language modelling constitutes its own task, learning how to handle negative polarity items such as \emph{any} or \emph{ever} in a grammatically correct way can be considered a subtask of language modelling.}
Examples of benchmarks that were originally meant to address this kind of cross-task transfer -- although they are not used as such any longer -- are multitask benchmarks such DecaNLP \citep{mccann2018natural}, GLUE \citep{wang-etal-2018-glue} and its successor SuperGLUE \citep{wang2019super}.
More recent benchmarks formulate all tasks as sequence-to-sequence problems \citep[e.g.][]{raffel2020t5, aribandi2022ext5, xie2022unifiedskg} so that they can be addressed with a single, typically very large, text-to-text language model.

\paragraph{The pretrain-finetune paradigm}
Cross-task generalisation is traditionally deemed an extremely challenging topic.
This has changed with the relatively recent trend of models that are first \emph{pretrained} with a general-purpose, self-supervised objective -- usually (masked) language modelling -- 
and then further \textit{finetuned} with the addition of task-specific parameters that learn to execute different tasks using the representations emerged in the pretraining phase.
The popularisation of this \emph{pretrain-finetune paradigm} has shifted thoughts on how to evaluate cross-task generalisation.
Rather than evaluating how learning one task can benefit another, this paradigm instead gives a central role to the question of how well a model that has acquired some general knowledge about language can successfully be adapted, with task-specific parameters, to different kinds of tasks \citep[e.g.][]{peters-etal-2018-deep,howard-ruder-2018-universal,devlin-etal-2019-bert,liu2019roberta}.
Interestingly, after finetuning, task performance is typically evaluated with random train--test splits, and thus generalisation \emph{within} individual tasks is not necessarily considered.\looseness-1

\paragraph{In-context learning}
In the most recent years, the focus of cross-task generalisation studies has shifted even further, to scenarios which consider how well pretrained language models fare in different tasks without the addition of task-specific parameters.
In the most extreme case, this implies evaluating a language model directly on a range of tasks without any further training.
To do so, tasks are reformulated as text-completion problems, such that language models can be \emph{prompted} directly with a question representing a specific task (\emph{zero-shot learning}), potentially preceded by a few examples (\emph{few-shot learning}) \citep[][]{radford2019language}.
The latter case, in which the intention is that models -- without any parameter updates -- `learn' from the examples given in the context, is often also referred to as \textit{in-context learning}.
Unfortunately, studies that investigate the relationship between the training and test data in such setups are rare, which leaves this young research area with many open questions.
A different class of in-context learning studies are those that finetune a pretrained model with prompts from one set of tasks, and then evaluate it on another set of tasks \citep[e.g.][]{zhong-etal-2021-adapting-language,wei2022finetuned,sanh2022multitask}.
While also in this case the pretraining corpus is uncontrolled, at least the relationship between the finetuning training and test data can be monitored, and the performances on the test data with and without finetuning easily compared; nevertheless, few studies do so.

\subsubsection{Generalisation across languages}
\label{subsec:crosslingual}

The fourth type of generalisation we include in our taxonomy is generalisation across languages, or cross-lingual generalisation.
Research in NLP has been very biased towards models and technologies for English \citep{bender2011achieving}
and most of the recent breakthroughs rely on amounts of data that are simply not available for the vast majority of the world's languages.
As well as from a practical perspective, work on cross-lingual generalisation is thus important for the promotion of inclusivity and democratisation of language technologies.
While the field of multilingual modelling is vast and naturally instigates interesting generalisation questions, relatively few papers in the area focus explicitly on cross-lingual generalisation.
In this section, we discuss two main strands of research that \textit{do} address this type of generalisation; in~\cref{appendix:multilingual_benchmarks}, we provide a list of benchmarks that can be used to evaluate generalisation across languages.

\paragraph{Cross-lingual finetuning}
There are several ways in which cross-lingual generalisation can be evaluated.
Most existing cross-lingual studies focus on scenarios where labelled training data is available in a single language (typically English) and the model is evaluated in multiple languages.
A common approach to address this problem is to finetune a multilingually pretrained language model on task-specific annotations available in one or a few languages, and then transfer to other languages in a zero-shot fashion \citep[e.g.][]{wu-dredze-2019-beto,pires-etal-2019-multilingual}. This setup tests to what extent a model's ability to solve tasks is invariant to the language of the labelled data used for training.
It has been used to show, for instance, that Multilingual BERT \citep{devlin-etal-2019-bert} finetuned on English labelled data generalises well to languages with different scripts, but exhibits some systematic deficiencies that affect language pairs with different word-order features, such as English and Japanese \citep{pires-etal-2019-multilingual}.

\paragraph{Multilingual learning}
A second way in which cross-lingual generalisation can be evaluated is by testing whether multilingual models perform better than monolingual models on language-specific tasks as a result of being trained on multiple languages at the same time.
As is the case for multitask learning, approaches that are simultaneously trained on multiple languages (or multiple tasks) can be thought of as an implicit evaluation of generalisation across those languages (or across tasks).
There is a large number of papers investigating multilingual models, usually for language modelling or machine translation \citep[e.g.][]{nllb2022no,aharoni-etal-2019-massively,zhang-etal-2020-improving,al-shedivat-parikh-2019-consistency,fan2021beyond}.
Most of these papers have as their main aim to introduce models that improve on multilingual tasks across the board and are not otherwise motivated by generalisation questions.
Some, however, do include explicit generalisation experiments in their setup, for example to assess the dependence of generalisation on the amount of data available for different languages \citep{zhou-etal-2018-massively}, or on the number of languages a model is exposed to during training \citep{aharoni-etal-2019-massively}.

\subsubsection{Generalisation across domains}
\label{subsec:domain}

The next category is generalisation across different domains, a type of generalisation that is often required in naturally occurring scenarios -- more so than the types discussed so far -- and thus carries high practical relevance.
While there is no precise definition of what constitutes a domain, domains broadly refer to collections of texts exhibiting different topical and/or stylistic properties, such as different genres or texts with varying formality levels.
Examples of domains we found in the literature are fiction, letters, governmental documents, telephone calls, and face-to-face interactions \citep{williams-etal-2018-broad}, biomedical texts \citep{fried-etal-2019-cross}, texts collected from online sources such as ArXiv, Github and OpenSubtitles \citep{artetxe2021moe}, or texts produced by different language communities, e.g.\ written in Standard American English and African-American English \citep{blodgett-etal-2016-demographic}.

\paragraph{Domain generalisation}
Studies considering domain generalisation are varied, and examples plentiful.
To name just a few, \citet{ryu-etal-2018-domain} and \citet{tan-etal-2019-domain} consider how a sentiment analysis model trained to classify the sentiment of reviews for certain products generalises to newly commercialised products, necessarily not represented in its training data.
\citet{blodgett-etal-2016-demographic} investigate how a model trained on data collected from one demographic generalises to a different population.
\citet{blodgett-etal-2017-dataset} and \citet{michel-neubig-2018-mtnt} instead investigate how a machine translation model trained on canonical text generalises to noisy data from an internet platform. 

\paragraph{Domain adaptation} Cross-domain generalisation has often been studied in connection with domain adaptation, the problem of adapting an existing general model to a new domain \citep{daume-iii-2007-frustratingly}.
This has been a very active research area in machine translation \citep{chu-wang-2018-survey,hu-etal-2019-domain-adaptation,axelrod-etal-2011-domain,luong-manning-2015-stanford,joty-etal-2015-avoid,wang-etal-2017-sentence,koehn-schroeder-2007-experiments,freitag2016fast,chu-etal-2017-empirical,bertoldi-federico-2009-domain,wang-etal-2017-instance}, with several standard datasets \citep{michel-neubig-2018-mtnt,malinin2021shifts} and dedicated tracks in popular shared tasks like WMT \citep{ws-2019-machine,specia-etal-2020-findings}.
It has also been studied in part-of-speech tagging \citep{blitzer-etal-2006-domain}, sentiment analysis \citep{blitzer-etal-2007-biographies} and language model pretraining \citep{gururangan-etal-2020-dont}, among other tasks.

\paragraph{Temporal generalisation} Finally, domain generalisation is related to temporal generalisation, as investigated in studies where the training data is produced in a specific time period and the model is tested on data from a different time period, either in the future or in the past.
This problem has as yet been studied in a limited range of tasks, including language modelling and question answering \citep{lazaridou2021mind}, named entity recognition in social media \citep{fromreide-etal-2014-crowdsourcing,rijhwani-preotiuc-pietro-2020-temporally,derczynski-etal-2016-broad}, named entity disambiguation \citep{agarwal-etal-2018-dianed}, document classification \citep{huang-paul-2018-examining,huang-paul-2019-neural-temporality,he2018time} and sentiment analysis \citep{lukes-sogaard-2018-sentiment}.\looseness-1

\subsubsection{Generalisation in the context of robustness}
\label{subsec:robustness}

The last category of generalisation research we consider on the generalisation type axis concerns models' ability to learn task solutions that abstract away from spurious correlations that may occur in the training data, and that are aligned with the underlying generalising solution that humans associate with the task \citep[e.g.][]{mccoy-etal-2019-right,talman-chatzikyriakidis-2019-testing,gururangan-etal-2018-annotation}.
We refer to this type of generalisation as \emph{robustness generalisation}.
Research on robustness generalisation usually focuses on data shifts that stem from varying data collection processes.
Different from most of the previous categories discussed in \cref{sec:generalisation_target}, such shifts are generally unintended and can be hard to spot.
Current work, therefore, focuses on characterising such scenarios and understanding their impact.
Many of these studies show that models do not generalise in the way we would expect them to, because the training data was in some subtle manner not representative of the true task distribution.
Generalisation evaluation in the context of robustness can be driven by several different motivations:
some studies are motivated by more practical concerns, others are conducted to gain a better perspective on intrinsic task understanding, and yet others are directed towards the development of fair and unbiased NLP models.
In this section, we discuss three common scenarios of robustness evaluation.

\paragraph{Annotation artefacts} A common scenario is one where there are annotation artefacts in the training data, which may result in an overestimation of a model's performance on a particular task.
Artefacts occur frequently when datasets are collected through crowdsourcing, with undesired data properties being introduced in subtle ways as a result of how the annotation procedure was set up.
Popular natural language inference datasets such as SNLI \citep{bowman-etal-2015-large} and MultiNLI \citep{williams-etal-2018-broad} have been found particularly susceptible to such artefacts.
For example, \citet{gururangan-etal-2018-annotation} and \citet{poliak-etal-2018-hypothesis} showed that models can learn to make correct predictions for NLI instances by only looking at hypotheses, with spurious patterns in word choice and grammatical features (e.g.\ negation being indicative of the \textit{contradiction} class) making it unnecessary for a model to use logical inference. The lack of true task understanding causes NLI models to generalise poorly across different datasets \citep{talman-chatzikyriakidis-2019-testing}.
Besides NLI, other tasks such as question answering have also been reported to suffer from annotation artefacts \citep{jia-liang-2017-adversarial,kaushik-lipton-2018-much}, even when the authors made a conscious effort to avoid such artefacts during the annotation process \citep{elazar-etal-2021-back}.\looseness-1

\paragraph{Standardised splits}
Another line of work questions the way we use data splits in general, especially the extent to which scores on standardised splits that remain static over time are reflective of a model's generalisation abilities.
For instance, \citet{gorman-bedrick-2019-need} showed that models perform much worse on fully random train--test splits than the reported state-of-the-art performances on standardised random splits.
\citet{sogaard-etal-2021-need} go even further and advocate for the use of heuristic and adversarial splits, thanks to which a model's capability for generalisation is challenged directly -- for instance by putting all longer sentences in the test set, or by splitting the data to maximise the difference between train and test set along a certain dimension.

\paragraph{Subpopulation bias} A third scenario in which robustness and performance overestimation play a role is the case where certain demographics are under- or over-represented in the training data.
As it may result in models that generalise poorly to specific demographic groups, this is a particularly harmful case of overestimation.
Toxicity classifiers, for example, suffer from unintended bias caused by certain identity terms being disproportionately represented in the training data \citep[e.g.\ \textit{``I am a gay man''} being assigned high toxicity scores because the word \textit{``gay''} is often used in toxic comments;][]{dixon2018measuring}, and abusive language detection models exhibit gender bias caused by imbalances in the training data \citep{park-etal-2018-reducing}.
A way to detect such imbalances and thus systematically avoid cases of overestimation is evaluating models by their worst-group accuracy, rather than the average accuracy across all demographic groups~\citep{koh2021wilds}. \looseness-1

\subsection{Shift type: what kind of data shift is considered?}
\label{sec:data_shifts}

\begin{figure}
    \centering
\includegraphics[width=\textwidth]{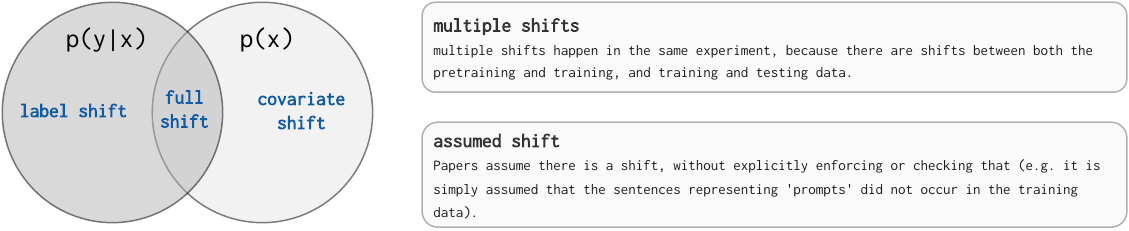}
    \caption{The five types of data distribution shifts (\textit{shift types}).}\label{fig:data_shifts}
\end{figure}

We have seen that generalisation tests may differ in terms of their motivation and the type of generalisation that they target.
What they share, instead, is that they all focus on cases in which there is a form of \emph{shift} between the data a model is (pre)trained on and the data that is used for evaluation.
In other words, for some datasets $(\datasetx_1,\datasety_1)$ and $(\datasetx_2,\datasety_2)$ considered in the experimental setup, it holds that $p(\bx_{1}, \by_{1}) \neq p(\bx_{2}, \by_{2})$.
In the third axis of our taxonomy, graphically depicted in \cref{fig:data_shifts}, we describe the ways in which two datasets used in a generalisation experiment can differ.
This axis adds a more formal dimension to our taxonomy and derives its importance from the fact that data shift plays an essential role in formally defining and understanding generalisation from a statistical perspective.

We consider three main types of shift which are well-attested in the literature -- \emph{covariate shift}, \emph{label shift} and \emph{full shift} -- and include two additional types of shift -- \textit{assumed shift} and \textit{multiple shifts} -- to account for studies that cannot be labelled with any of the three main shift types.
%
We formalise the differences between the test, training and potentially pretraining data involved in generalisation tests as shifts between the respective \emph{data distributions}:
\begin{align}
    &p(\xtest, \ytest)&\qquad \textcolor{gray}{\texttt{test}} \label{eq:test_data_distribution} \\
    &p(\xtrain, \ytrain)&\qquad \textcolor{gray}{\texttt{training / finetuning / adaptation}}  \label{eq:training_data_distribution}\\
    &p(\xpretrain, \ypretrain)&\qquad \textcolor{gray}{\texttt{pretraining}} \label{eq:pretraining_data_distribution}
\end{align}
These data distributions can be expressed as the product of the probability of the input data $p(\bx)$ and the conditional probability of the output labels given the input data $p(\by|\bx)$:
\begin{align}
p(\bx, \by) &= p(\bx)\  p(\by|\bx) \label{eq:types-of-shift-train}
\end{align}
This allows us to define four main types of relations between two data distributions, depending on whether the distributions differ in terms of $p(\mathbf{x})$, $p(\by|\bx)$, both, or none.
The last type constitutes the case in which there is no shift in data distributions -- i.e. both $p(\xtrain)=p(\xtest)$ and $p(\ytrain|\xtrain)=p(\ytest|\xtest)$.\footnote{For clarity, we leave pretraining distributions aside and focus on train--test shifts, as this is the most intuitive setting. However, the shift types described in this section can be used to describe the relationship between any two data distributions involved in a modelling pipeline.}
This matches the i.i.d.\ evaluation setup traditionally used in machine learning.
As discussed earlier, this type of evaluation, also referred to as evaluation of \emph{within-distribution} generalisation, has frequently been reported not to be indicative of good performance for the more complex forms of generalisation that we often desire from our models.
We will not further discuss it here, but instead focus on the other three cases, commonly referred to as \emph{out-of-distribution} (o.o.d.) evaluation.
\Cref{fig:data_shifts} summarises the types of distribution shift discussed in this section.

\subsubsection{Covariate shift}
The most commonly considered data distribution shift in o.o.d.\ generalisation research is one where $p(\xtest)\!\neq\!p(\xtrain)$ but $p(\ytest|\xtest)\!=\!p(\ytrain|\xtrain)$.
In this scenario, often referred to as \emph{covariate shift} \citep{storkey2009training,moreno2012unifying}, the distribution of the input data $p(\bx$) changes, but the conditional probability of the labels given the input -- which describes the task -- remains the same.
Under this type of shift, one can evaluate if a model has learned the underlying task distribution while only being exposed to $p(\xtrain, \ytrain)$.
Challenge sets such as HANS~\citep{mccoy-etal-2019-right}, PAWS~\citep{yang-etal-2019-paws}, or the COGS test set~\citep{kim-linzen-2020-cogs} deliberately address these shifts, with examples being selected or generated to violate invalid heuristics models are known or expected to follow.
Covariate shift is also addressed in cross-generalisation and robustness evaluation studies, such as those conducted by \citet{ryu-etal-2018-domain} and \citet{tan-etal-2019-domain} on real-world datasets.
Compared to the other shift types, covariate shift is the easiest to tackle without performing additional training or pre- and post-processing.
As we will see in what follows, a common approach to address other, more complex shifts, is to turn them into covariate shifts.

\subsubsection{Label shift}
The second type of shift corresponds to the case in which the focus is
on the conditional output distributions: $p(\xtest)\!=p(\xtrain)$ and $p(\ytest|\xtest)\!\not=\!p(\ytrain|\xtrain)$. We refer to this case as \emph{label shift} but it is also known as \textit{concept shift} in the literature~\citep{moreno2012unifying}.
Label shift can happen within the same task when there are inter-annotator disagreements, a temporal shift in the data, or a change of domain (e.g.\ the phrase ``it doesn't run'' can lead to different sentiment labels depending on whether it appears in a review for software or one for mascara).
Label shift also occurs when there is a change in task (as in \cref{subsec:crosstask}).
For instance, the same sentence might have a negative gold label in a sentiment classification task, but a positive label when the task is changed to toxicity identification.
Or, in case of a more extreme label shift, the labels themselves can change, for example when shifting from language modelling (where the set of labels is the language vocabulary) to POS-tagging.
In NLP studies, label shift is often seen as an obstacle that needs to be overcome rather than as a setting in which models are directly evaluated: if the same example has contradictory labels in training and test data, it is unclear what decision at test time should be considered good generalisation behaviour.
In practice, there are two main ways in which label shift is typically addressed.
The first is to add a finetuning or adaptation stage in which a model is updated to represent the shift that occurred \citep[e.g.][]{sun2020lamol,biesialska-etal-2020-continual} or new parameters are added to represent newly introduced labels \citep[i.a.]{howard-ruder-2018-universal,peters-etal-2018-deep,devlin-etal-2019-bert}.
The second way to address label shift is to augment the input data with domain or task indicators \citep[e.g.][]{brown2020language,raffel2020t5}.
We saw before that the phrase ``it doesn't run'' can be both positive and negative, depending on its domain of occurrence.
By adding indicators that specify the domain,
the problem can be converted into a covariate shift (or potentially even no shift, if both indicators are represented in the training and test distributions).  
Similarly, in prompting setups, where tasks are formulated as questions in natural language, label shifts caused by a change of task are turned into a different shift type that can be solved without further finetuning \citep[see, e.g.][]{brown2020language,schick-schutze-2021-exploiting,bach-etal-2022-promptsource}.

\subsubsection{Full shift}
The most extreme type of shift corresponds to the case in which $p(\bx)$ and $p(\by|\bx)$ change simultaneously: $p(\xtest) \not= p(\xtrain)$ and $p(\ytest|\xtest) \not= p(\ytrain|\xtrain)$.
We refer to this case as \emph{full shift}.
Full shifts may occur in language modelling tasks, where changes in the $p(\bx)$ directly translate into changes in $p(\by|\bx)$,
when adapting to new language pairs in multi-lingual experiments \citep[e.g.][]{nllb2022no,kodner-etal-2022-sigmorphon},
or when entirely different types of data are used either for pretraining \citep[e.g.][who test if pretraining on music impacts learning language afterwards]{papadimitriou-jurafsky-2020-learning} or for evaluation \citep[e.g.][who evaluate generalisation to different languages]{de-varda-zamparelli-2022-multilingualism}.
Full shifts can be addressed without retraining -- because they do not necessarily imply that the same input $x$ is assigned a different label at test time.
Nevertheless, they are challenging, and, similarly to label shifts, they are often turned into different types of shifts that can be more easily addressed.%
\footnote{Oftentimes, covariate shifts might inadvertently also cause label shifts, for instance when the textual domain changes in a sequence-classification task.
In our characterisation, however, if the underlying task stays the same, we will assume that the (more controlled) covariate shift is the one that is investigated unless specified otherwise.}

\subsubsection{Multiple shifts}
We have so far focused on the types of shifts that can occur between two data distributions.
Some studies, however, consider shifts between multiple distributions at the same time,
for instance to investigate how different types of pretraining architectures generalise to o.o.d.\ splits in a finetuning stage \citep{li-etal-2022-quantifying} or which pretraining method achieves better cross-domain generalisation in a second training stage \citep{wang-etal-2021-meta}.
In our taxonomy, we label such cases as \emph{multiple shifts}, and -- at least in the current version -- we do not distinguish between different configurations of multiple shifts (e.g.\ label+covariate, or covariate+covariate).\footnote{Our \emph{evaluation cards} (\cref{sec:eval_cards}), however, do allow recording different shift types per experiment.}
We will discuss multiple shifts further in \cref{sec:shift_locus}.

\subsubsection{Assumed shift}
When classifying shifts in our review, we will mainly focus on cases where authors explicitly consider the relationship between the data distributions they use in their experiments and the assumptions they make about this relationship are either well-grounded in the literature (e.g.\ it is commonly assumed that switching between domains constitutes a covariate shift) or empirically verified.
Nevertheless, we identify numerous studies that claim to be about generalisation where such considerations are absent: it is assumed that there is a shift between training and test data, but this is not verified or grounded in previous research. We include this body of work in our review and refer to the corresponding type of shift as \emph{assumed shift}.
Sometimes, the assumed shift is not explicitly checked because it is considered plausible given general linguistic knowledge \citep[e.g.][]{wilcox-etal-2021-targeted}.
Other times, the relationship between training and test data is not investigated because the researchers do not have access to the training data.
The BigBench benchmark \citep{srivastava2022beyond}, for instance, contains several tasks designed to measure generalisation, but the training datasets of the models investigated are not in the public domain.
Yet in other cases, the training data is available to the authors of the paper, but no extensive analysis is presented \citep[e.g.][]{chowdhery2022palm,brown2020language}.

\subsection{Shift source: how are training and test data obtained?}
\label{sec:shift_sources}

\begin{figure}
	\centering
	\includegraphics[width=1.0\textwidth,trim=0mm 0mm 0mm 0mm, clip]{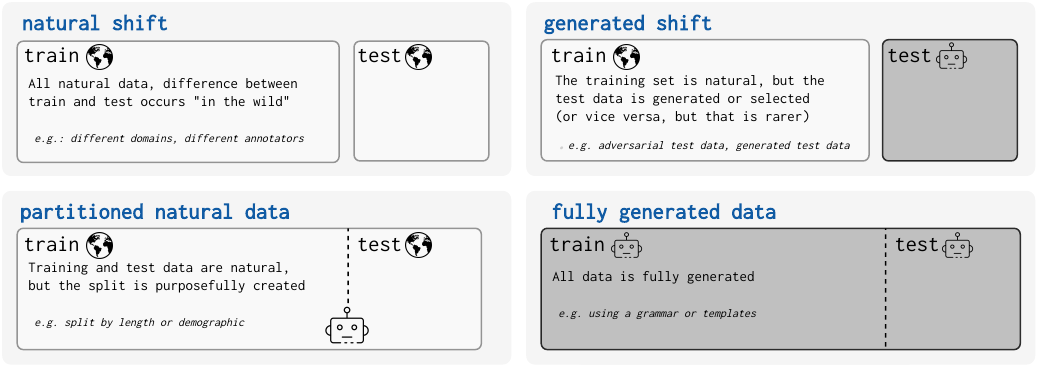}
	\caption{The sources of shifts, with indications of whether data is natural (globe) or generated (robot).\looseness-1}\label{fig:shift_source}
\end{figure}

In the previous section, we discussed what types of shifts may occur in generalisation tests.
We now focus on how those shifts originated: our fourth axis, graphically shown in \cref{fig:shift_source},
concerns the \emph{source} of the differences occurring between the pretraining, training and test data distributions.
The source of the data shift determines how much control an experimenter has over training and test data and, consequently, what kind of conclusions can be drawn from a generalisation experiment.
We distinguish four different sources of shifts:~(i)~\emph{naturally occurring shifts},
(ii)~\emph{artificially partitioned natural corpora},
(iii)~\emph{generated shifts}
and (iv)~\emph{fully generated datasets}.

To formalise the description of these different sources of shift, we consider the unobserved \textit{base distribution} which describes all data considered in an experiment:\looseness=-1%
\begin{align}
	&p(\xbase, \ybase, \btau)&\qquad\qquad \textcolor{gray}{\texttt{base}} \label{eq:task_data_distribution}
\end{align}
The variable $\btau$ represents a \emph{data property of interest}, with respect to which a specific generalisation ability is tested.
This can be an observable property of the data (e.g.\ the length of an input sentence), an unobservable property (e.g.\ the timestamp that defines when a data point was produced), or even a property relative to the model under investigation (e.g.\ $\btau$ could represent how quickly a data point was learned in relation to overall model convergence).
The base distribution over $\bx$, $\by$ and $\btau$ can be used to define different partition schemes to be adopted in generalisation experiments.
Formally, a partitioning scheme is a rule $f:\!\mathcal{T}\!\rightarrow\!\{\texttt{true, false}\}$ that discriminates data points according to a property $\btau\!\in\!\mathcal{T}$.
To investigate how a partitioning scheme impacts model behaviour, the pretraining, training and test  distributions can be defined as:
\begin{align}
	p(\xpretrain, \ypretrain) &= p(\xbase, \ybase\!\mid\!f_{\texttt{pretrain}}(\btau) = \texttt{true}) \\
	p(\xtrain, \ytrain) &= p(\xbase, \ybase\!\mid\!f_{\texttt{train}}(\btau) = \texttt{true})\\
	p(\xtest, \ytest) &= p(\xbase, \ybase\!\mid\!f_{\texttt{test}}(\btau) = \texttt{true})
\end{align}
Using these data descriptions, we can now discuss four different sources of shifts.

\subsubsection{Naturally occurring shifts}
The first scenario we consider is one in which shifts naturally occur between corpora.
Both the data partitions of interest are naturally occurring corpora, to which no systematic operations are applied: for the purposes of a generalisation test, experimenters have no direct control over the base distribution nor the partitioning scheme~$f(\btau)$. In other words, the variable $\btau$ refers to properties that naturally differ between collected datasets.
Examples of naturally occurring shifts emerge from splits containing data from different annotators \citep{geva-etal-2019-modeling}, sources or domains \citep[e.g.][]{artetxe2021moe,talman-chatzikyriakidis-2019-testing}, populations \citep[e.g][]{dixon2018measuring,talat2018bridging}, time periods \citep[e.g.][]{lazaridou2021mind}, or from different data collection procedures targeting the same task \citep{williams-etal-2018-broad,wang-etal-2018-glue}.
In this category, we also include cross-task and cross-lingual generalisation studies in which all corpora involved are natural corpora \citep[e.g.][]{mishra-etal-2022-cross,fitzgerald2022massive}.\looseness-1

\subsubsection{Splits of natural corpora}
A slightly less natural setup is one in which a naturally occurring corpus is used, but it is artificially split along specific dimensions.
The primary difference with the previous category is that the variable $\btau$ refers to properties along which data would not naturally be split, such as the length or syntactic complexity of a sample.
Experimenters have thus no control over the data itself, but they control the partitioning scheme~$f(\btau)$.
\citet{raunak-etal-2020-long}, for instance, split naturally occurring machine translation corpora such that longer sentences occur in the test data, and \citet{weber-etal-2021-language} split a language modelling corpus such that the training data does not contain specific types of grammatical environments. 

\subsubsection{Generated shifts}
The third category concerns cases in which one data partition is a fully natural corpus and the other partition is designed with specific properties in mind to address a generalisation aspect of interest.
Data in the constructed partition may avoid or contain specific patterns \citep{fancellu-etal-2017-detecting,bhargava-etal-2021-generalization,cui-etal-2022-generalized-quantifiers,dankers-etal-2022-paradox}, violate certain heuristics \citep{mccoy-etal-2019-right,dayanik-pado-2021-disentangling,libovicky-etal-2022-dont}, include unusually long or complex sequences \citep{lakretz2021causal,raunak2019compositionality}, or it may be constructed \emph{adversarially}, generated either by humans \citep{kiela-etal-2021-dynabench} or automatically \citep[e.g.][]{zellers-etal-2018-swag,sakaguchi2021winogrande}.
In the examples provided above,
the constructed partition always corresponds to the test data; the opposite -- where instead the \emph{training data} is synthetic or generated and the test data natural --~is also possible, yet less common \citep[e.g.][]{papadimitriou-jurafsky-2020-learning}.

\subsubsection{Fully generated}
The last possibility is to use only generated data.
Generating data is often the most precise way of measuring specific aspects of generalisation as experimenters have direct control over both the base distribution and the partitioning scheme.
Sometimes the data involved is entirely synthetic \citep[e.g.][]{hupkes2020compositionality,lake2018generalization}, other times it is templated natural language or a very narrow selection of a natural language corpus \citep[e.g][]{kim-linzen-2020-cogs,keysers2019measuring}.
Generated splits can vary in several different dimensions.
Sometimes, $\btau$ is a simple observable data property.
For instance, \citet{hupkes2020compositionality} split their corpus based on the presence of particular function pairs $\mathcal{P}$, implicitly setting $\btau\!=\!\mathcal{P}\in x$.
In some cases, $\btau$ may also be defined relative to the $\btau$ of other examples, and can only be computed globally, such as in the case of \textit{maximum compound divergence} splitting \citep{keysers2019measuring}.\looseness-1\footnote{Maximum compound divergence is not restricted to generated data, but can in some cases also be applied to natural data.}

\subsection{Locus of shift: between which data distributions does the shift occur?}
\label{sec:shift_locus}

\begin{figure}
\centering
\includegraphics[width=1.0\textwidth,trim=0mm 0mm 0mm 0mm, clip]{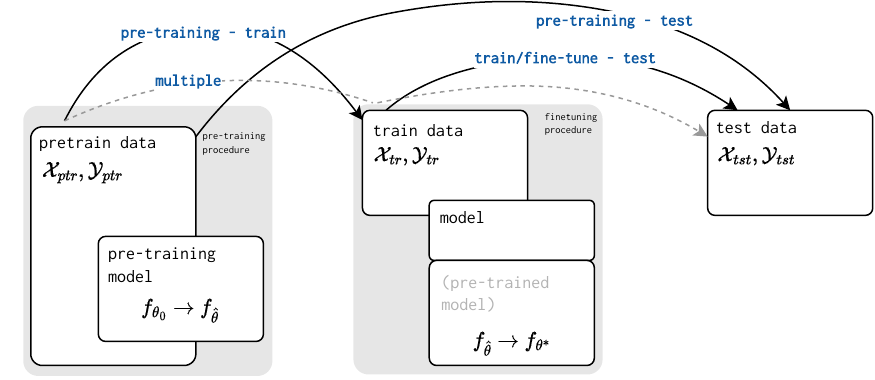}
\caption{The five loci of splits, along with the parts of the modelling pipeline they allow investigating.}\label{fig:shift_locus}
\end{figure}

The four axes that we have discussed so far demonstrate the depth and breadth of generalisation evaluation research, and they also clearly illustrate that generalisation is evaluated in a wide range of different experimental setups.
They described the high-level motivations for studying generalisation in NLP models, the types of generalisation that have been frequently evaluated in the literature, the data distribution shifts used for generalisation tests, and the possible sources of those shifts.
What we have not yet explicitly discussed is between which data distributions those shifts can occur: the \textit{locus} of the shift.
In our taxonomy, the shift locus forms the last piece of the puzzle, as it determines what part of the modelling pipeline is investigated and, with that, what kind of generalisation questions can be answered.
We consider shifts between all stages in the contemporary modelling pipeline -- pretraining, training and testing, as well as studies that consider shifts between multiple stages at the same time, as expressed by the data distributions that we have considered in \cref{sec:data_shifts} (for a graphical representation, we refer to \cref{fig:shift_locus}).
Given these distributions, there exist five possible loci of shifts: shifts between the \emph{training and test data}, between the \emph{finetuning training and test data}, between the \emph{pretraining and finetuning training data}, between the \emph{pretraining and test data}, and between \emph{all data distributions}.

We describe the five loci of shift and how they interact with different components of the modelling pipeline with the aid of three \emph{modelling distributions}.
These modelling distributions correspond to the previously described stages -- testing a model, training it, and potentially pretraining it:
\begin{align}
	&p(\datasetytest \mid \datasetxtest, \thetaopt)&\qquad \textcolor{gray}{\texttt{model}} \label{eq:model_distribution} \\
	&p(\thetaopt \mid \datasetxtrain, \datasetytrain, \bphi_{tr}, \thetainitial)&\quad \textcolor{gray}{\texttt{training/finetuning/adaptation}} \label{eq:training_distribution}  \\
	&p(\thetainitial\!\mid \datasetxpretrain, \datasetypretrain, \bphi_{pr}, \btheta_0)&\textcolor{gray}{\texttt{pretraining}} \label{eq:pretraining_distribution}
\end{align}
In these equations, $\bphi$ broadly denotes the training and pretraining hyperparameters, $\btheta$ refers to the model parameters, and $\datasetx, \datasety$ indicate sets of inputs and their corresponding output.
\Cref{eq:model_distribution} defines a model instance, specifying a probability distribution over the target test labels $\datasetytest$ given the model's parameters $\thetaopt$ and a set of test inputs $\datasetxtest$.
\Cref{eq:training_distribution} defines a training procedure, specifying a probability distribution over model parameters $\thetaopt \in \R^d$ given a training dataset $\datasetxtrain$, $\datasetytrain$, a set of training hyperparameters $\bphi_{tr}$, and a (potentially pretrained) model initialisation $\thetainitial$.
Lastly, \cref{eq:pretraining_distribution} defines a pretraining procedure, specifying a conditional probability over the set of parameters $\thetainitial$,
given a pretraining dataset, a set of pretraining hyperparameters $\bphi_{pr}$, and a model initialisation.\footnote{Note that this formalisation generalises to the \textit{training from scratch} paradigm when $\datasetxpretrain, \datasetypretrain\!=\!\emptyset, \emptyset$, and to the \emph{in-context-learning} setup when $\datasetxtrain, \datasetytrain\!=\!\emptyset, \emptyset$.}
Between which of these stages a shift occurs impacts which modelling distributions can be evaluated.
We now discuss the different potential loci of shifts.

\subsubsection{The train--test locus}
Probably the most commonly occurring locus of shift in generalisation experiments is the one between training and test data, corresponding to the classic setup where a model is trained on some partition of the data and then directly evaluated on a shifted (out-of-distribution) test partition.
Studies with the train--test locus can assess two different parts of the modelling pipeline.
In some cases, researchers investigate the generalisation abilities of a \textit{model instance}.
Studies of this type, therefore, report the evaluation of a single set of parameters $\thetaopt$ as described in \cref{eq:model_distribution} -- typically made available by others -- without considering how exactly it was trained and how that impacted the model's generalisation behaviour.
For example, a surge of studies considered the behaviour of the pretrained language model made available by \citet{gulordava-etal-2018-colorless}, to investigate how it generalises to, for instance, different syntactic constructions \citep[e.g.][]{lakretz-etal-2019-emergence}.\footnote{
    The investigation of model instances is, however, more common with the \textit{pretrain-test} locus that we will discuss later in this section.}
Alternatively, researchers might evaluate one or more training procedures, investigating if the \textit{training distribution} results in model instances that generalise well -- for example, to study how generalisation compares between different architectures \citep{mul2019siamese,saxton2019analysing} or how it is affected by the amount of training data \citep[e.g.][]{artetxe2021moe,rae2021gopher}.
While these cases also require evaluating model instances, the focus of the evaluation is not on one particular instance, but rather on the procedure that generated the (multiple) evaluated model instances.\looseness-1

\subsubsection{The finetune train--test locus}
The second potential locus of shift bears similarities to the first one but instead considers data shifts between the train and test data used \textit{during finetuning}, and thus concerns models that have gone through an earlier stage of training.
This locus occurs when a model is evaluated on a finetuning test set that contains a shift with respect to the finetuning training data
\citep{kavumba-etal-2022-prompt,damonte-monti-2021-one,ludwig-etal-2022-improving}.
Studies with a finetune train--test locus can evaluate the same parts of the modelling pipeline as studies with a train--test locus. However, studies that investigate the generalisation abilities of individual finetuned model instances are rare.
More frequently, research with this locus focuses on the finetuning procedure and on whether it results in finetuned model instances that generalise well on the test set.
Experiments evaluating o.o.d.\ splits during finetuning often also include a comparison between different pretraining procedures (e.g.\ they compare how BERT models and RoBERTa models behave during finetuning),
thus investigating both a pretrain--train shift and a finetune train--test shift. We will mark them as having \emph{multiple loci}, as will be further discussed in the last subsection. \looseness-1

\subsubsection{The pretrain-train locus}
A third possible locus of shift is between pretraining and training data.
Experiments with this locus evaluate whether a particular pretraining procedure, as described in \cref{eq:pretraining_distribution}, results in models (parameter sets $\thetainitial$) that are useful when further trained on different tasks or domains.
For instance, \citet{artetxe2021moe} investigate which pretraining procedure shows the best downstream generalisation in several different tasks, \citet{tian-etal-2021-diagnosing} investigate how well pretrained models generalise to a newly proposed first-order-logic dataset, and \citet{freitag2016fast} test how well a pretrained NMT model can adapt to different domains.
Crucially, we classify studies as having a pretrain-train locus only when i.i.d.\ splits are used in their final training stage. 
If also the final stage contains a shift, we describe the study as having \textit{multiple loci}.

\subsubsection{The pretrain--test locus}
The fourth locus of shift is between pretraining and test data.
This locus occurs when a pretrained model is evaluated directly on o.o.d.\ data, without further training (i.e.\ $\datasetxtrain, \datasetytrain = \emptyset, \emptyset$) -- as frequently happens in in-context learning setups \citep[e.g.][]{li2021xglm,zhang2022opt}  -- or when a pretrained model is finetuned on examples that are i.i.d.\ with respect to the pretraining data and then tested on out-of-distribution instances.
The former case ($\thetaopt\!=\!\thetainitial$) is similar to studies with only one training stage in the train--test locus, but distinguishes itself by the nature of the (pre)training procedure, which typically has a general purpose objective, rather than being task-specific (e.g.\ a language modelling objective).
Furthermore, while generalisation studies with a train--test locus almost always explicitly consider the relationship between training and test data, this is frequently not the case with pretrain--test studies, where data shifts are assumed.

\subsubsection{Multiple loci}
The last option on the locus axis describes studies which simultaneously investigate multiple shifts between different parts of the modelling pipeline. We refer to these cases as generalisation tests with \emph{multiple loci}.
More explicitly, experiments of this type consider shifts both between the pretraining and the training data, as well as between the training and the test data.\footnote{
We do not distinguish cases where the test data is shifted with respect to the pretraining data from cases where it is not, as the latter are very uncommon. It is, however, possible to set up an experiment where the pretraining and test data are drawn from the same distribution, for example to test whether a finetuning procedure results in catastrophic forgetting.}
Multiple-loci experiments evaluate all stages of the modelling pipeline at once: they assess the generalisability of models produced by the pretraining procedure as well as whether generalisation is achieved in the finetuning stage \citep[e.g.][]{tu-etal-2020-empirical,hu2020xtreme,yanaka-etal-2021-exploring,fitzgerald2022massive}.
Because multiple-loci experiments necessarily also contain multiple shifts, we mark them as \emph{multiple shifts} in the shift type axis.
The nature of the two shifts may not be the same, but
in our analysis, we group them all into a single category.
In our proposed evaluation cards (\cref{sec:eval_cards}), however, different loci within a single experiment can be recorded separately.

\section{A review of existing generalisation research}
\label{sec:survey}

We presented a taxonomy containing five categorical axes that can be used to characterise generalisation research.
We now use the taxonomy to analyse a large amount of existing generalisation research and create a comprehensive map indicating which areas are covered and which are still unexplored.
More specifically, we consider \Nentries generalisation experiments in NLP, presented in a total of \Npapers papers from the ACL Anthology that have the (sub)words \emph{generalisation}, \emph{generalization}, \emph{generalise} or \emph{generalize} in their title or abstract, and we label them with their axis values on the five taxonomy axes.
In \cref{appendix:setup}, we provide more details on the selection procedure of the papers.
The full list of papers is provided in \hyperlink{appendix:references}{Appendix F}, as well as -- in searchable form -- on our website.\footnote{The full list of papers reviewed, on our website: \url{https://genbench.github.io/references}}
On the same website, we furthermore present \href{https://genbench.org/visualisation}{interactive ways} to visualise the results; a \href{https://genbench.org/references}{search tool} to retrieve relevant citations; and a means to \href{https://genbench.org/eval_cards}{generate \emph{evaluation cards}}, that authors can put in their paper or appendix to comprehensively summarise which generalisation experiments they did (for an example, we refer to \autoref{fig:eval_card} and \autoref{sec:eval_cards}).
In this section, we present the main findings of our analysis.

\begin{figure}
    \centering
    \begin{subfigure}[b]{0.26\textwidth}
        \includegraphics[width=\textwidth]{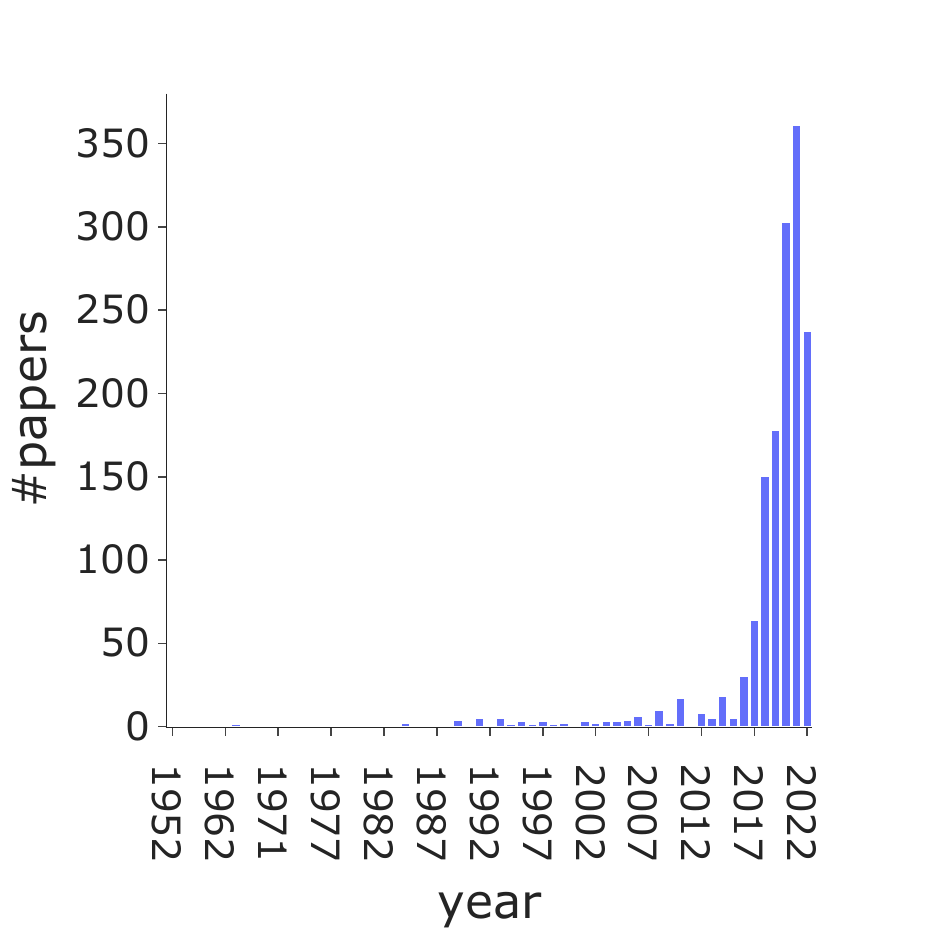}
        \caption{}\label{fig:generalisation_over_time_a}
    \end{subfigure}
   \begin{subfigure}[b]{0.26\textwidth}
       \includegraphics[width=\textwidth]{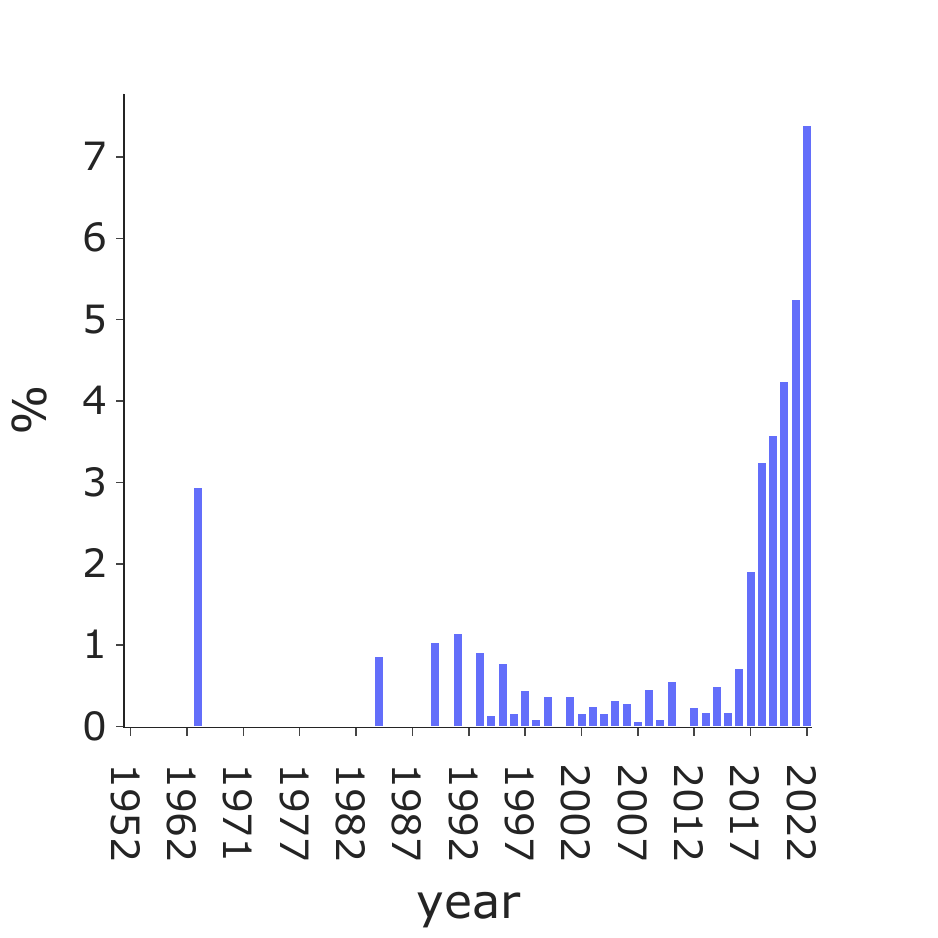}
       \caption{}\label{fig:generalisation_over_time_b}
   \end{subfigure}
    \begin{subfigure}[b]{0.42\textwidth}
        \includegraphics[width=\textwidth]{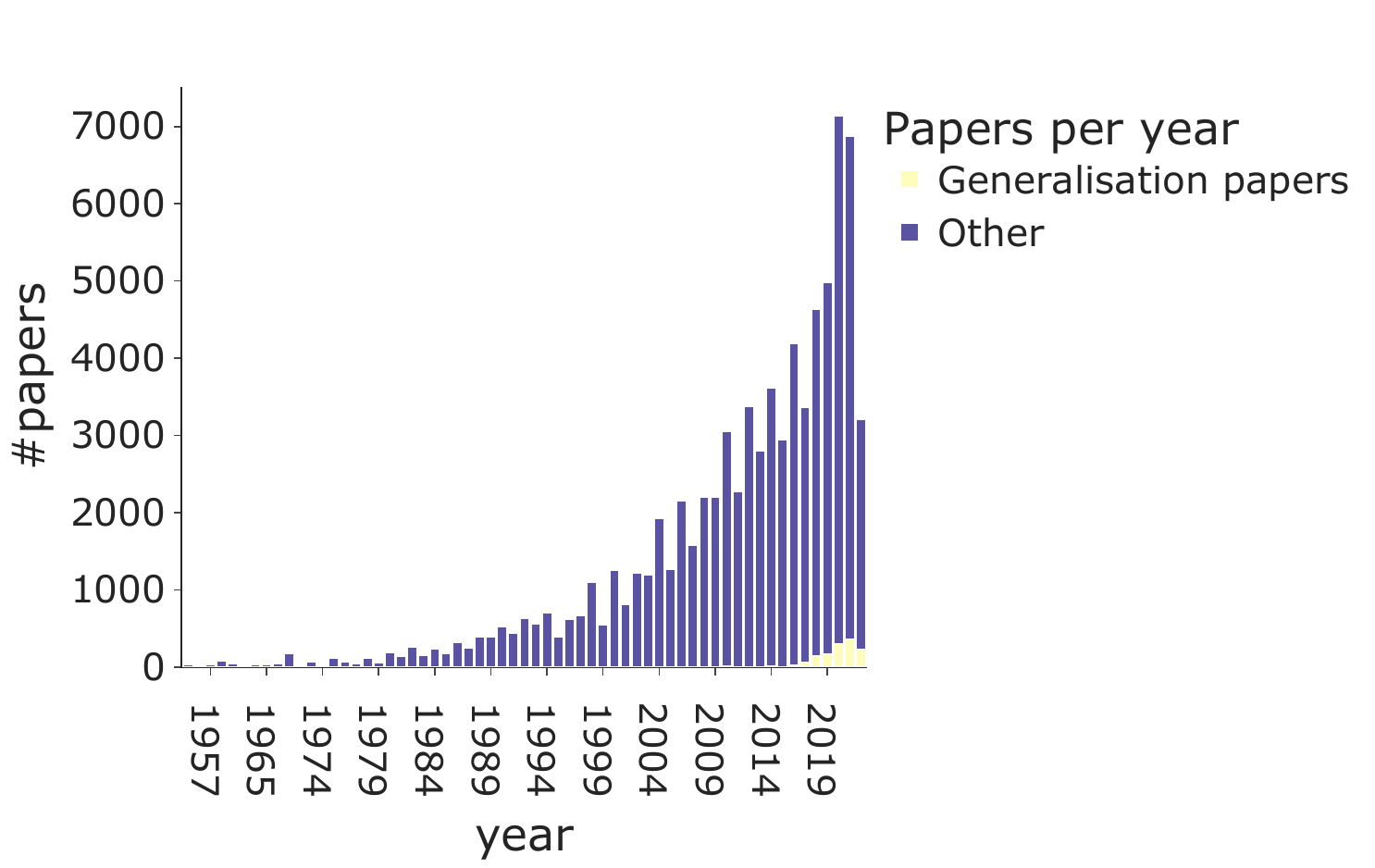}
        \caption{}\label{fig:generalisation_over_time_c}
    \end{subfigure}
    \caption{We selected papers from the ACL Anthology that contain the (sub)words  \emph{generalisation}, \emph{generalization}, \emph{generalise} or \emph{generalize} in their title or abstract. This figure shows how many such papers exist per year, both absolutely (a) and percentually (b). In (c), we show the number of generalisation papers published each year together with the total number of papers per year.}\label{fig:generalisation_over_time}
\end{figure}


\subsection{Overall trends on different axes}
\label{subsec:axis-trends}

\begin{figure}[!t]
\centering
\includegraphics[width=0.65\textwidth]{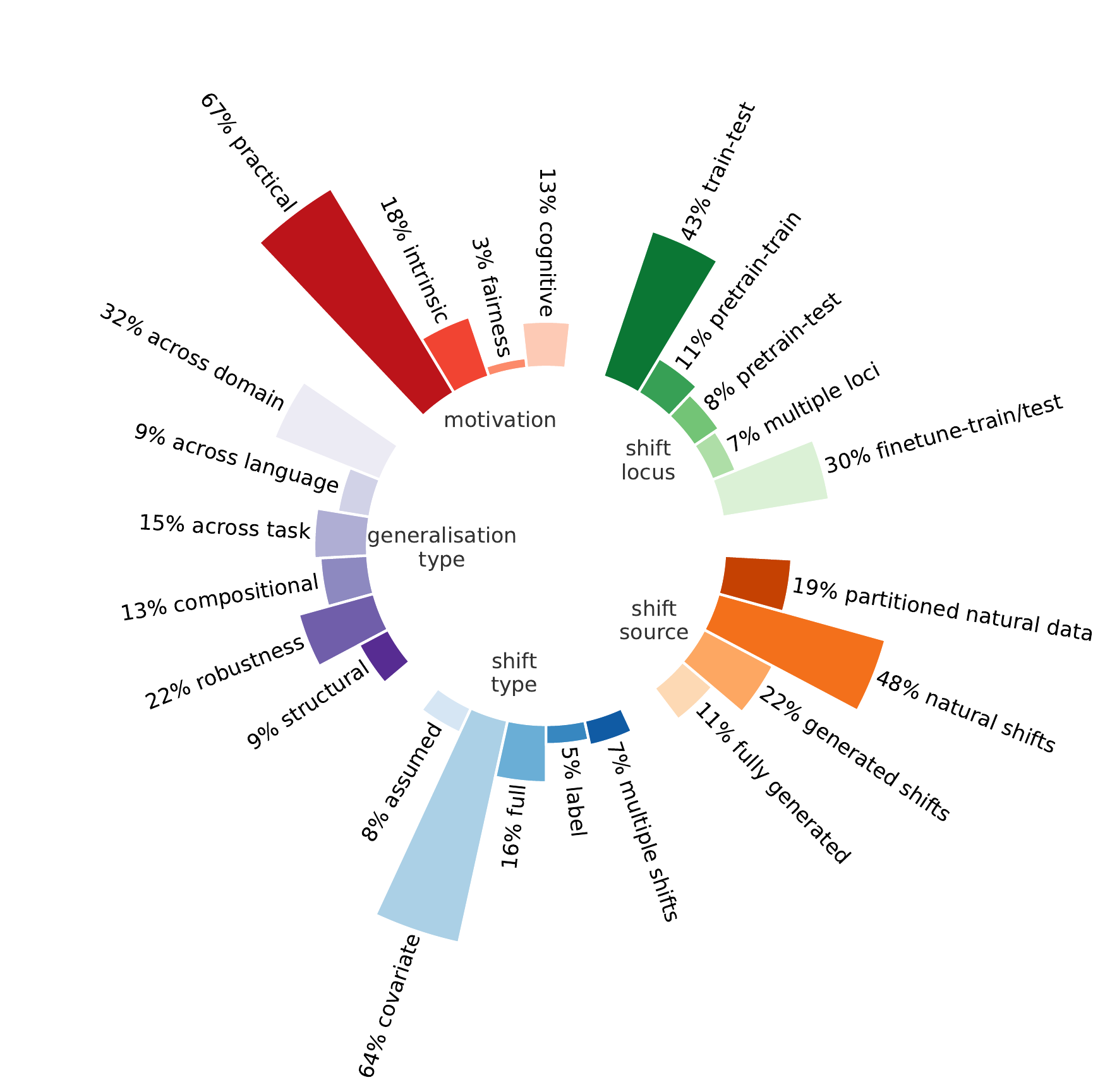}
\caption{The relative occurrences of the categories in the five different axes of our taxonomy (shown clockwise are the motivation, the generalisation type, the shift source, the shift type and the shift locus).}\label{fig:axes_value_counts}
\end{figure}

\begin{figure}
    \centering
\begin{subfigure}{0.32\textwidth}
    \centering
\includegraphics[width=1.0\textwidth]{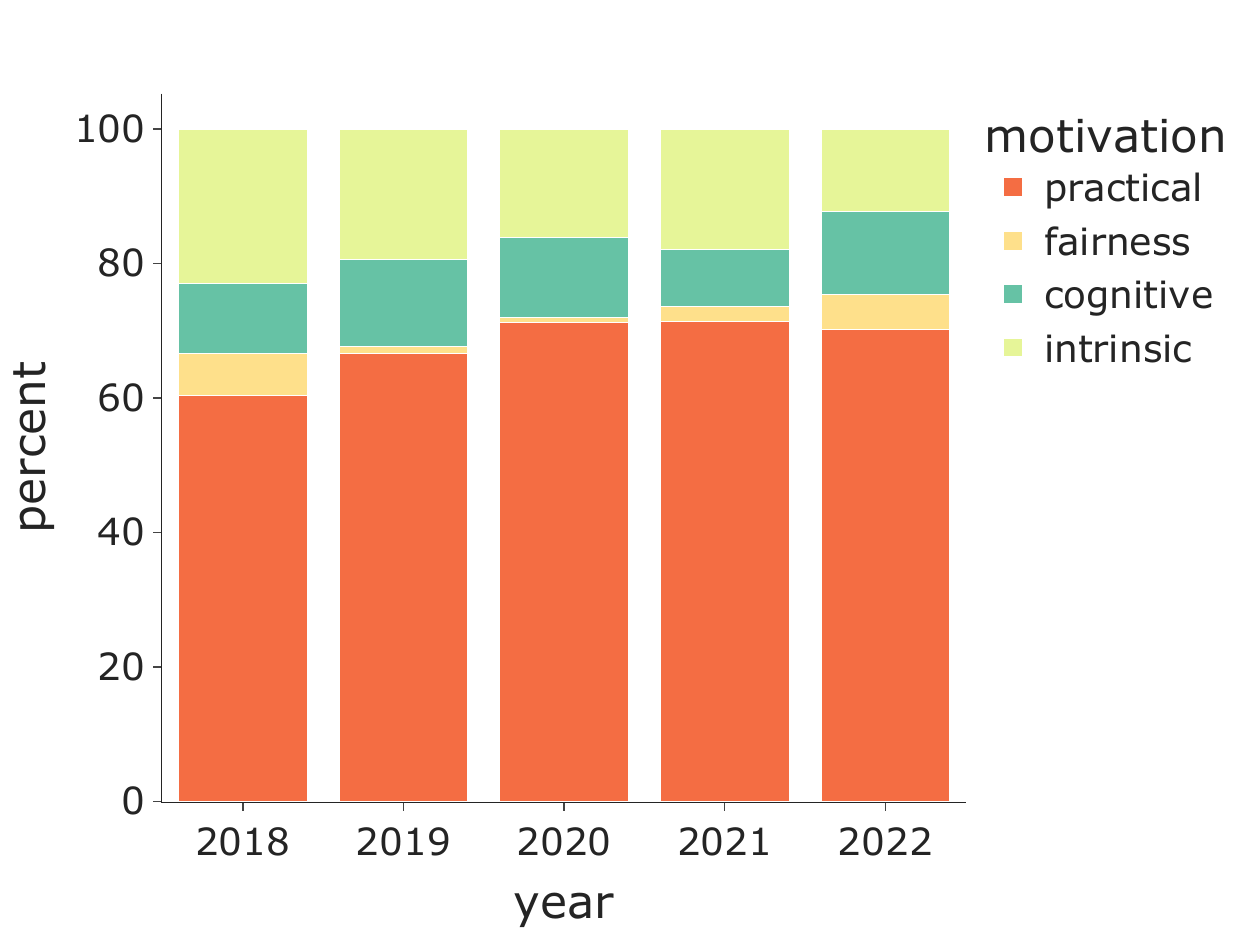}
    \caption{Motivation}\label{subfig:motivation-over-time-norm}
\end{subfigure}
\hfill
\begin{subfigure}{0.32\textwidth}
    \centering
\includegraphics[width=\textwidth]{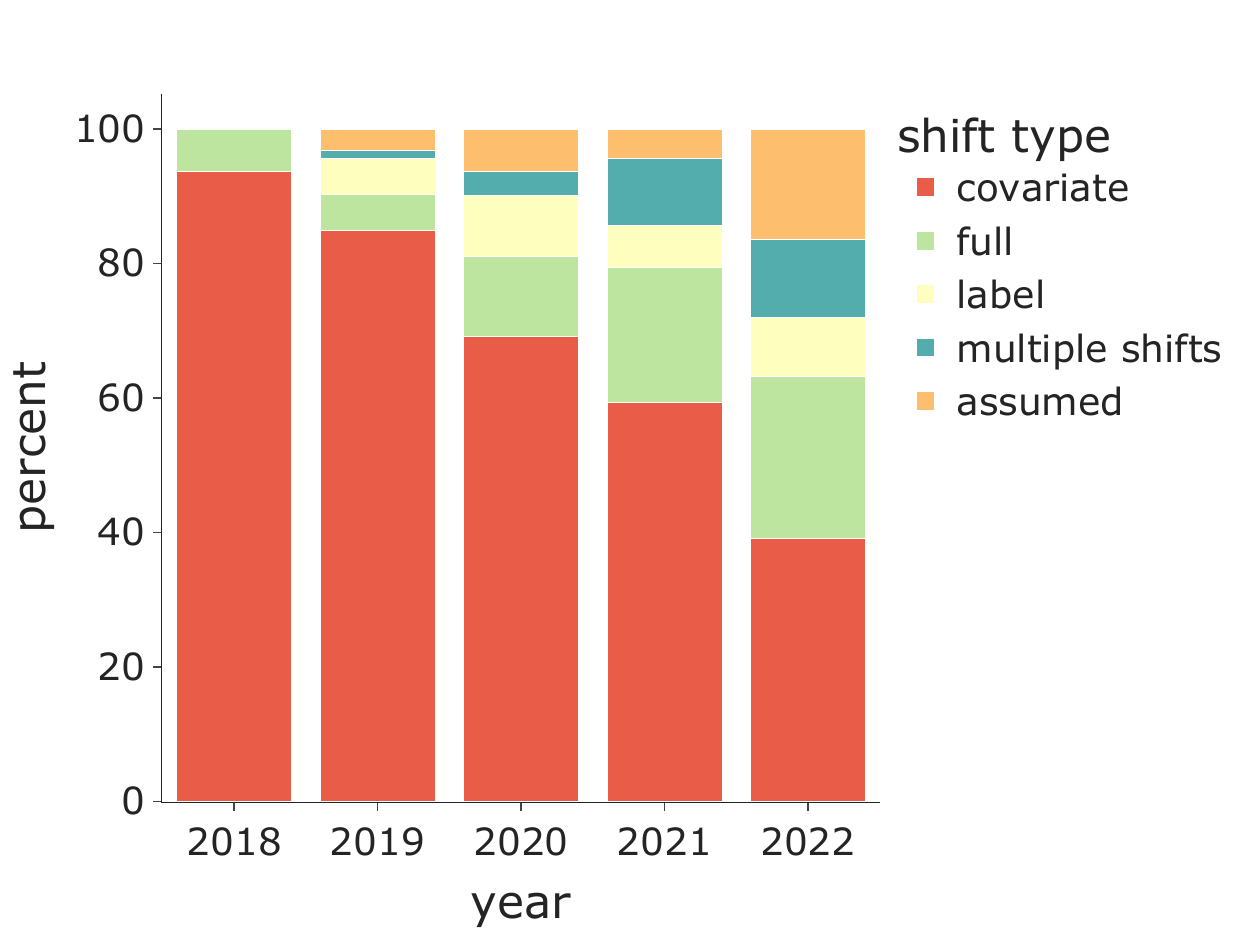}
    \caption{Shift type}\label{subfig:shift-over-time-norm}
\end{subfigure}
\hfill
\begin{subfigure}{0.32\textwidth}
    \centering
\includegraphics[width=\textwidth]{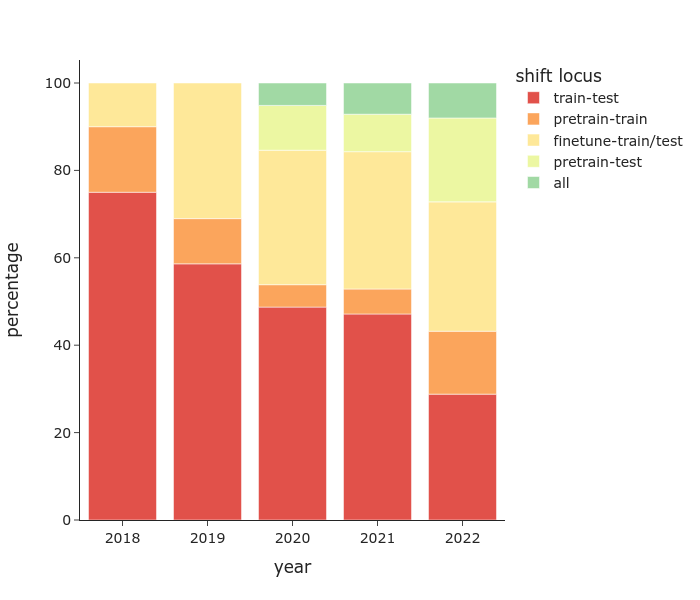}
    \caption{Shift locus}\label{subfig:locus-over-time-norm}
\end{subfigure}
    \caption{Trends from the past five years for three of the taxonomy axes (motivation, shift type and shift locus), normalised by the total number of papers annotated per year.}\label{fig:axes-over-time}
\end{figure}

We begin by discussing the overall frequency of occurrence of different categories on the five axes, without taking into account interactions between them.
We plot the relative frequencies of all axis values in \cref{fig:axes_value_counts} and their development over time in \cref{fig:axes-over-time}.
Because the number of generalisation papers retrieved before 2018 is very low (see \cref{fig:generalisation_over_time_a}), we restrict the diachronic plots to the last five years; all other statistics reported are computed over our entire selection of papers.

\subsubsection{Motivations}
As we can see in \cref{fig:axes_value_counts} (top left), by far the most common motivation to test generalisation is the practical motivation.
The intrinsic and cognitive motivations follow whereas the studies in our review that consider generalisation from a fairness perspective make up only 3\% of the total.
In part, this low number could stem from the fact that our keywords search in the anthology (see \cref{appendix:setup} for more information) was not optimal for detecting fairness studies, and we welcome researchers to submit other generalisation studies with a fairness motivation for review.
However, we also speculate that only relatively recently attention is starting to grow for the potential harmfulness of models trained on large, uncontrolled corpora and that generalisation has as yet simply not been studied extensively in the context of fairness.
Overall, we see that trends on the motivation axis have experienced small fluctuations over the past five years (\cref{subfig:motivation-over-time-norm}) but they have remained relatively stable.

\subsubsection{Generalisation type}
We find that cross-domain is the most frequent generalisation type, making up more than 30\% of all studies, followed by robustness, cross-task and compositional generalisation (\cref{fig:axes_value_counts}, left side).
Structural and cross-lingual generalisation are the least commonly investigated.
On the one hand, studies investigating structural generalisation may be underrepresented as they typically focus more on whether models can capture structures at all, rather than on whether they generalise to new structures.
On the other hand, while cross-lingual studies may be undersampled as they tend to less frequently use the word `generalisation' in their title or abstract (sometimes in favour of `transfer'), we hypothesise that their low number is reflective of the English-centric disposition of the field.
As for fairness studies, we encourage researchers to suggest cross-lingual generalisation papers that we may have missed via our website so that we can better estimate to what extent cross-lingual generalisation is in fact understudied.

\subsubsection{Shift type}
Data shift types (\cref{fig:axes_value_counts}, bottom) are very unevenly distributed over their potential axis values: the vast majority of generalisation research considers covariate shift.
Given that covariate shift is more easily addressed by most current modelling techniques, and that it can occur between any two stages of the modelling pipeline -- while label and full shift typically occur between pretraining and finetuning -- this is, to some extent, to be expected.
More unexpected, perhaps, is the relatively high amount of \emph{assumed} shifts, which correspond to studies that claim to test generalisation but do not explicitly consider how the test data relates to data used at various stages of model training.
The percentage of assumed shifts has in fact increased over the past few years (\cref{subfig:shift-over-time-norm}).
We hypothesise that this trend, which signals a movement of the field in the wrong direction, is predominantly caused by the use of increasingly large, general-purpose training corpora.
Such large corpora, which are often also not in the public domain, make it very challenging to analyse the relationship between the training and testing data and, consequently, to determine what kind of conclusions can be drawn from evaluation results.
More promising, instead, is the fact that several studies consider \emph{multiple shifts}, thus assessing generalisation throughout the entire modelling pipeline.

\subsubsection{Shift source}
On the shift source axis (\cref{fig:axes_value_counts}, bottom right), we see that almost half of the reviewed generalisation studies consider naturally occurring shifts, natural corpora that are not deliberately split along a particular dimension.
As discussed later in the current section, this type of data source is most prevalent in cross-task and cross-domain generalisation studies, for which such naturally different corpora are widely available.
The next most frequent categories are generated shifts, where one of the datasets involved is generated with a specific generalisation property in mind, and artificially partitioned natural data, describing settings in which all data is natural, but the way it is split between train and test is controlled.
Fully generated datasets are less common, making up only 11\% of the total number of studies.

\subsubsection{Shift locus}
Lastly, for the locus axis (\cref{fig:axes_value_counts}, top right), we see that the majority of cases focuses on (finetune) train--test splits.
Much fewer studies focus on shifts between pretraining and training or pretraining and testing.
Similar to the previous axis, a comparatively small percentage of studies considers shifts in multiple stages of the modelling pipeline.
At least in part, this might be driven by the larger amount of compute that is required for those scenarios.
Over the last five years (\cref{subfig:locus-over-time-norm}), however, the percentage of studies considering multiple loci and pretrain--test loci -- the two least frequent categories -- has increased.\looseness-1

\subsection{Interactions between axes}
\label{subsec:interactions}
Next, we consider interactions between different axes.
Are there any combinations of axes that occur together very often or combinations that are instead rare?
We encourage the reader to explore these interactions dynamically on our website.
Here, we discuss a few relevant trends.

\begin{figure}
    \centering
\begin{subfigure}[b]{0.49\textwidth}
    \centering
    \includegraphics[width=0.8\textwidth]{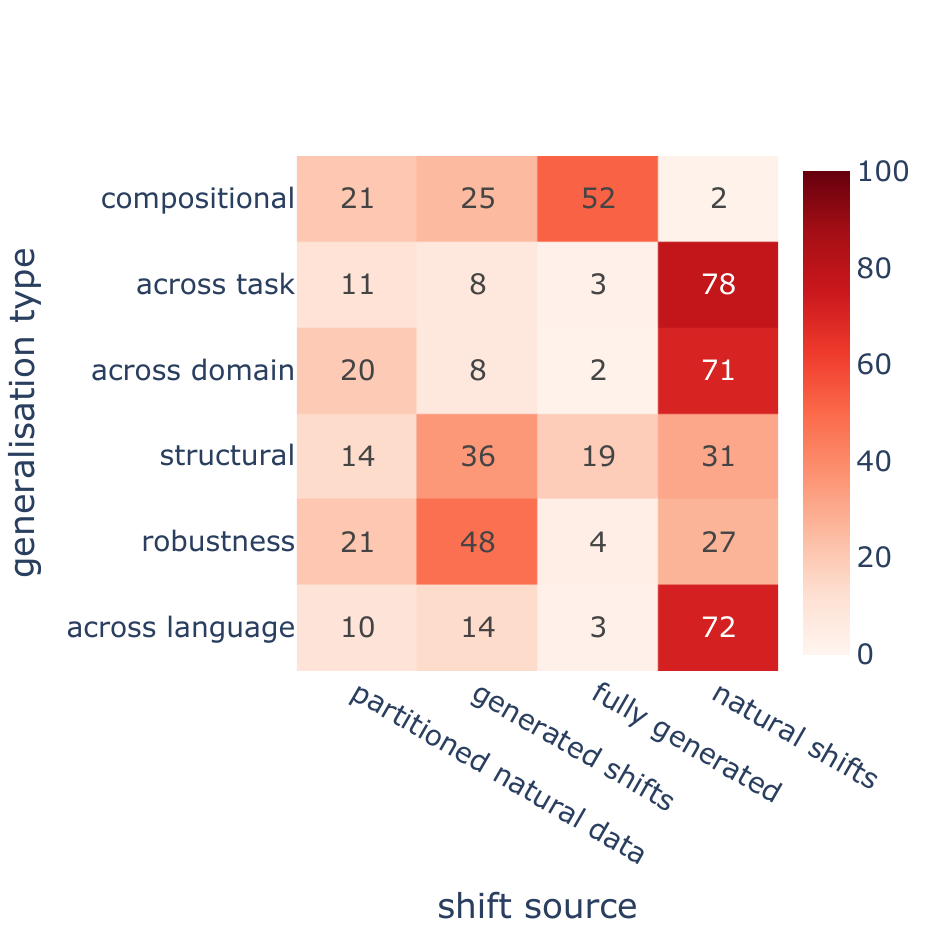}
    \caption{Data source per generalisation type}\label{subfig:source-per-type}
\end{subfigure}\hfill
\begin{subfigure}[b]{0.49\textwidth}
    \centering
    \includegraphics[width=0.8\textwidth]{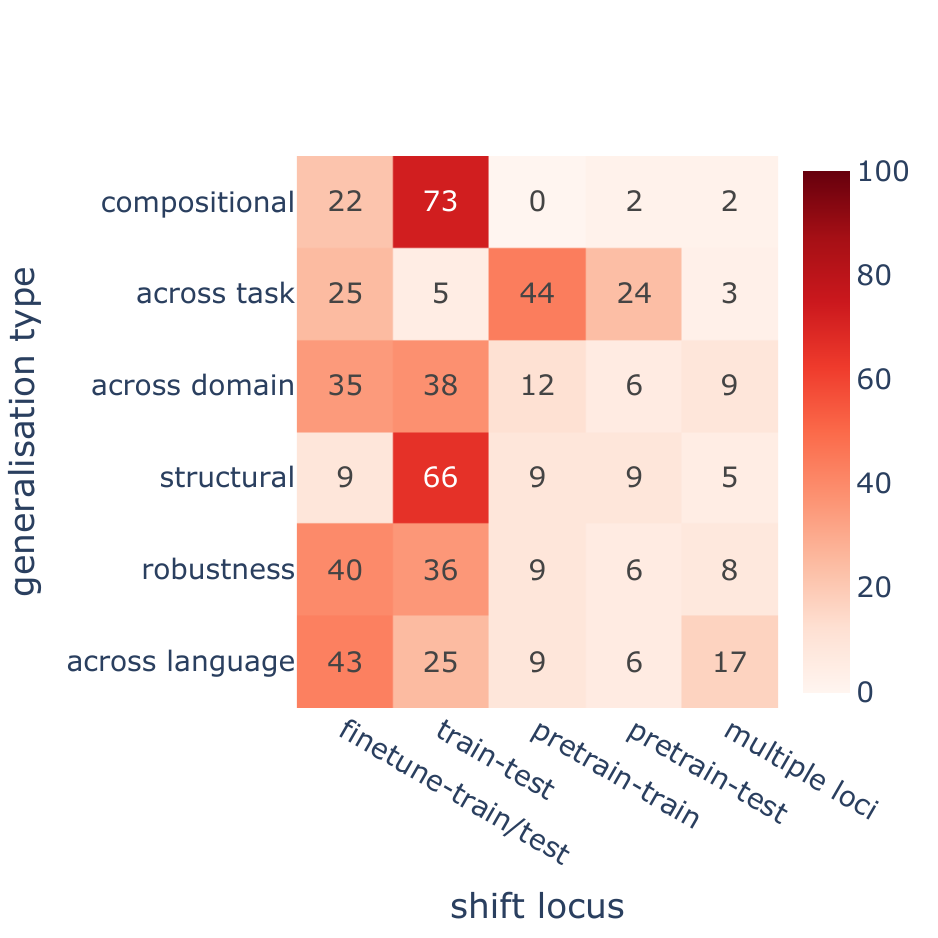}
    \caption{Shift locus per generalisation type}\label{subfig:locus-per-type}
\end{subfigure}

\begin{subfigure}[b]{0.49\textwidth}
    \centering
    \includegraphics[width=0.8\textwidth]{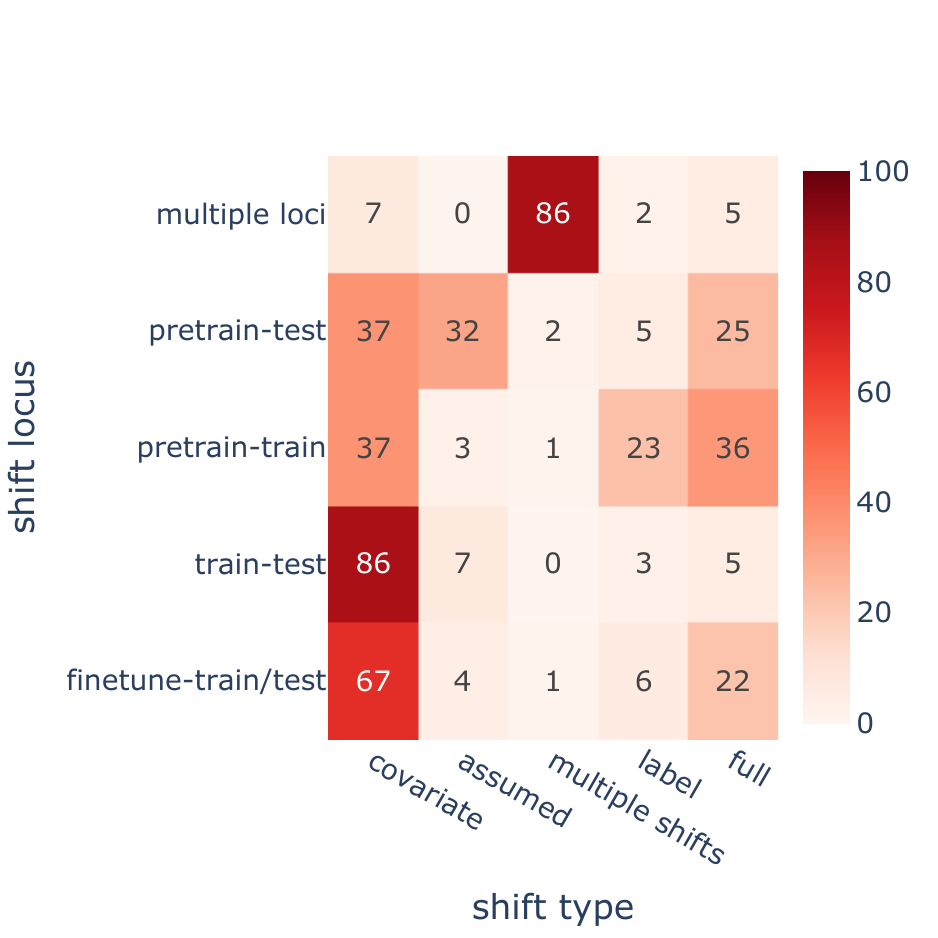}
    \caption{Shift type per locus}\label{subfig:shift-per-locus}
\end{subfigure}\hfill
\begin{subfigure}[b]{0.49\textwidth}
    \centering
    \includegraphics[width=0.8\textwidth]{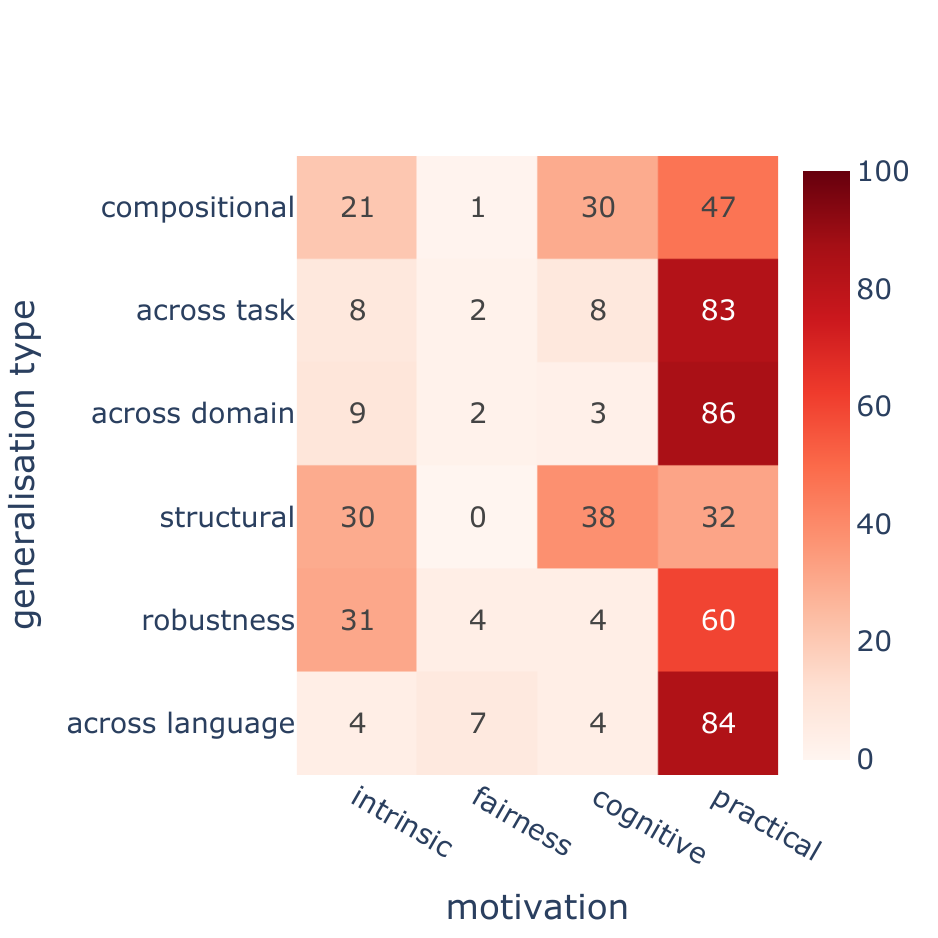}
    \caption{Motivation per generalisation type}\label{subfig:motivation-per-type}
\end{subfigure}

\begin{subfigure}[b]{0.49\textwidth}
    \centering
    \includegraphics[width=0.8\textwidth]{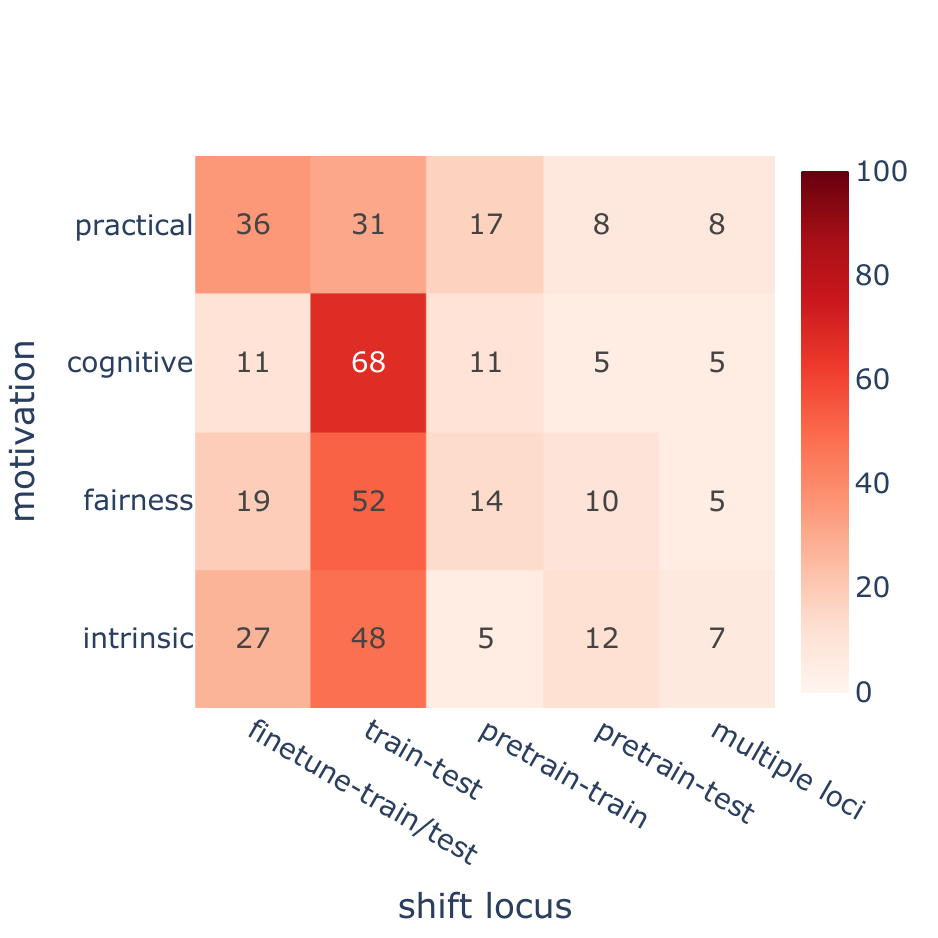}
    \caption{Locus per motivation}\label{subfig:locus-per-motivation}
\end{subfigure}\hfill
\begin{subfigure}[b]{0.49\textwidth}
    \centering
    \includegraphics[width=0.8\textwidth]{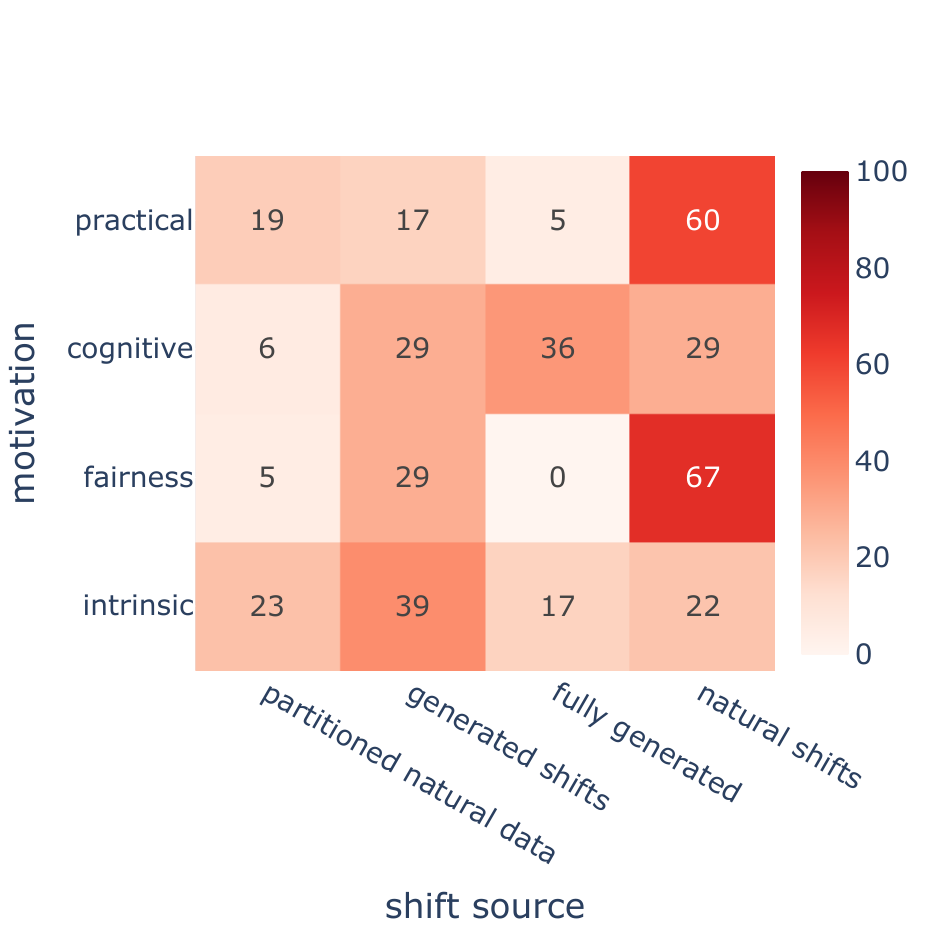}
    \caption{Shift source per generalisation type}\label{subfig:source-per-motivation}
\end{subfigure}

    \caption{Interactions between axes. The heatmaps are normalised by the total row value to facilitate comparisons between rows.
    Different normalisations (e.g.\ to compare columns) and interactions between other axes can be analysed on our website, where figures based on the same underlying data can be generated.}\label{sec:interactions}
\end{figure}

\subsubsection{What data shift source is used for different generalisation types?}
In \cref{subfig:source-per-type}, we show the frequency of each data source per generalisation type, normalised by the total number of times that generalisation type occurs (i.e.\ row sum, to make patterns comparable between generalisation types).
The shift source varies widely across different types of generalisation.
Compositional generalisation, for instance, is predominantly tested with fully generated data, a data type that hardly occurs in research considering robustness, cross-lingual or cross-task generalisation.
Those three types of generalisation are most frequently tested with naturally occurring shifts or, in some cases, with artificial splits of natural corpora.
Structural generalisation is the only generalisation type that appears to be tested across all different data types.
As far as we are aware, there exist very few studies that directly compare results between different sources of shift -- for instance, to investigate to what extent results on generated shifts or fully generated data are indicative of performances on natural corpora.\footnote{An example of such a study would be the work of \citet{chaabouni-etal-2021-transformers}, who investigate whether performance improvements on SCAN transfer to machine translation models trained on natural data.}
Such studies could provide insight into how choices in the experimental design impact the conclusions that are drawn from generalisation experiments, and we believe that they are an important direction for future work.

\subsubsection{For which loci of shift are different generalisation types studied?}
We observe that of all the generalisation types, only cross-task generalisation is frequently investigated in the pretrain-train and pretrain--test stages (\cref{subfig:locus-per-type}).
For all other types of generalisation, the vast majority of tests are conducted in the train--test or finetune-train/test stage.
In some cases, these differences are to be expected: as general-purpose pretrained models are usually trained on very large, relatively uncontrolled corpora, investigating how they generalise to a different domain without further finetuning is typically not possible, and neither is evaluating their robustness, which typically requires more detailed knowledge of the training data.
The statistics also confirm the absence of studies that consider compositional generalisation from pretraining to finetuning or from pretraining to training, which as we previously discussed (\cref{subsec:compositional}) is philosophically and theoretically challenging in such setups.
A final observation is the relative underrepresentation of studies with multiple loci across all generalisation types, especially given the large number of studies that consider generalisation in the finetuning stage or the pretrain-training stage.
Those studies have included multiple training stages but considered generalisation in only one of them.
We hope to see this trend change in the future, with more studies considering generalisation across the entire modelling pipeline.

\subsubsection{Which types of data shifts occur across different loci?}
\cref{subfig:shift-per-locus} shows that assumed shifts mostly occur in the pretrain--test locus, which confirms our hypothesis that assumed shifts are likely caused by the use of increasingly large, general-purpose training corpora.
When such pretrained models are further finetuned, they often consider either a shift between pretraining and finetuning where new labels are introduced or a covariate shift in the finetuning stage -- as such, they do not require an in-depth understanding of the pretraining corpus.\footnote{The observant reader might note that there are, in fact, also several covariate and full shifts with a pretrain-train locus, as well as covariate shifts with a pretrain--test locus. These typically do not represent experiments with large language models but instead, for instance, consider a multi-stage process for domain adaptation, which also includes a zero-shot comparison.}
When models are directly evaluated, however, the only shift that can be considered is the one between the very large pretraining corpus and the test corpus.
This trend points to a substantial challenge when it comes to evaluating generalisation with limited knowledge about model pretraining.

\subsubsection{How does motivation drive generalisation research?}
To discuss the relationship between the motivation behind a study and the other axes, we focus in particular on its interactions with generalisation type, shift locus and shift source, as shown in \cref{subfig:motivation-per-type}-\ref{subfig:source-per-motivation}.
Considering first the relationship between motivation and generalisation type (\cref{subfig:motivation-per-type}), we see that cross-domain, robustness, cross-task and cross-lingual generalisation are predominantly motivated by practical considerations, with robustness generalisation studies being also frequently motivated by an interest in understanding how models work intrinsically.
We find that compositional and structural generalisation studies are both frequently driven by cognitive motivations -- which is to be expected given the importance of these concepts in human cognition and intelligence.
The motivation given most frequently for compositional generalisation, however, is a practical one.
While in human learning, compositionality is indeed often associated with important practical properties -- speed of learning, quick generalisation -- as far as we know, there is little empirical evidence that compositional models actually perform better on natural language tasks.
A similar apparent mismatch can be observed in \cref{subfig:source-per-motivation} when focusing on the practical motivation.
Practical generalisation tests are typically aimed at improving models or at being directly informative of a model's applicability.
Nonetheless, more than 20\% of the practically motivated studies use either artificially partitioned natural data or even fully generated data.
To what extent could their conclusions then actually be informative of models applied in practical scenarios?
These apparent mismatches between the motivation and the experimental setup demonstrate the importance of the motivation axis in our taxonomy -- being aware of and explicit about a study's motivation ensures that its conclusions are indeed informative to the underlying research question.

Another interesting observation that can be made from the interactions between motivation and shift locus is that the vast majority of cognitively motivated studies are conducted in a train--test setup.
While there are many good reasons for this, conclusions about human generalisation are drawn from a much more varied range of `experimental setups'.
For instance, any experiments done with adults can be thought of as more similar to tests with finetune train--test or pretrain--test locus than to the train--test locus, as adults have a life-long experience over which the experimenter has little control beyond participant selection.
On the one hand, this suggests that generalisation with a cognitive motivation should perhaps be evaluated more often with those loci.
On the other hand, it begs the question of whether the field could take inspiration from experiments on human generalisation for the challenging effort of evaluating the generalisation of large language models, trained on uncontrolled corpora, in a pretrain--test setting.
While there are, of course, substantial differences between the assumptions that can reasonably be made about the linguistic experiences of a human and the pretraining of a language model,\footnote{
On the one hand, for a human, some assumptions can be safely made or even verified with a participant -- for instance, unless a person has previously participated in a psycholinguistic experiment, we can almost be certain that they have never conjugated \emph{nonce words}.
For a computational model, this is less trivially true.  
On the other hand, it is sometimes possible to inspect the data that models have seen during pretraining, which is evidently not the case for humans.} we believe that input from domain experts who have extensively studied generalisation in humans might be very beneficial to improving model generalisation testing in these more challenging setups.\looseness-1


\section{Conclusion}
\label{sec:conclusion}

While the ability to generalise well 
is considered to be one of the primary desiderata for NLP models, there is very little agreement on what kind of generalisation behaviour modern-age NLP models should exhibit, and under what conditions that should be evaluated.
For decades, generalisation has been simply evaluated with random train--test splits.
The recent past, however, has seen several studies illustrating that models that exhibit near-perfect performances on such i.i.d.\ splits can sometimes drastically fail in a wide range of scenarios that require different forms of generalisation.
This body of work demonstrates the need for more comprehensive generalisation testing but it does not provide much guidance: different papers use different experimental setups, different types of data and even entertain different ideas about what it means for an NLP model to generalise well.
As a consequence, even though its importance is almost undisputed, extensive state-of-the-art generalisation testing is not currently the standard in NLP.
With our work, we aim to set the first step towards making it the new status quo.
In this last section, we summarise our work, discuss its limitations, and sketch how we believe it can be used in the future.

\subsection{The generalisation taxonomy}
We presented a new framework to systematise and understand generalisation research.
The core of this framework consists in a taxonomy that characterises generalisation studies along five dimensions.
This taxonomy, which is designed based on an extensive review of generalisation papers in NLP, can be used to critically analyse existing generalisation research and to structure new studies.
The five nominal axes of the taxonomy describe why a study is executed (the main \textbf{motivation} of the study), what the study intends to evaluate (the \textbf{type} of generalisation it aims to solve), and how the evaluation is conducted (the type of \textbf{data shift} considered, the \textbf{source} of this data shift, and the \textbf{locus} in which the shift is investigated).
An overview of our taxonomy is provided in Figure~\ref{fig:taxonomy-infographic}; the axes are discussed in \cref{sec:motivations}-\ref{sec:shift_locus}.
For the reader's convenience, a concise summary is provided in \cref{appendix:taxonomy_summary}.

\subsection{Existing work on generalisation: the taxonomy in action }
To illustrate the use and usefulness of our taxonomy, we used it to analyse \Npapers papers from the ACL Anthology that have the (sub)words generali(s/z)ation or generali(s/z)e in their title or abstract.
Through this analysis, we demonstrated that the taxonomy is applicable to a wide range of generalisation studies and we were able to provide a comprehensive map of the field, observing overall patterns and making suggestions for areas that should be prioritised in the future.
Our most important conclusions and recommendations are: \looseness-1
\begin{itemize}\setlength\itemsep{0.01em}
    \item The goal of a study is not always perfectly aligned with its experimental design. We advise that future work should be more explicit about motivations -- which strongly impact what sort of generalisation is even desirable -- and should incorporate deliberate assessments to ensure that the experimental setup matches the goal of the study. To facilitate this, we introduce \emph{evaluation cards} (see \autoref{sec:eval_cards}) that can be used to comprehensively report which generalisation experiments are conducted in a paper.
    \item Cross-lingual studies and generalisation studies motivated by fairness goals are underrepresented. We suggest that these areas should be given more attention in future work.
    \item Papers that target similar generalisation questions vary widely in the type of evaluation setup they use. In our view, the field would benefit from more \emph{meta-studies} that consider how the results of experiments with different experimental paradigms compare to each other.
    \item The vast majority of generalisation studies focuses on only one stage of the modelling pipeline. More work is needed that considers generalisation in all stages of training, to prioritise models whose generalising behaviour persists throughout their training curriculum.
    \item Recent popular NLP models that can be tested directly for their generalisation from pretraining to testing (e.g.\ in prompting setups, without any further model training) have often been evaluated without considering the relationship between the (pre)training and the test data. We envisage that this is due to the fact that generalisation is particularly difficult to assess when large uncontrolled training data is involved, and we suggest that inspiration might be taken from how generalisation is evaluated in experiments with human participants, where control and access to the `pretraining' data of a participant are unattainable.
\end{itemize}

While the review and conclusions presented in this paper are necessarily static, along with this paper we also launch a \href{https://genbench.github.io}{website}, on which new entries can be added by authors.
On this website, we furthermore provide a set of visualisation tools that make it possible to visualise our results in different ways and a set of search tools that allows browsing through the reviewed papers, finding studies with specific features, and collecting relevant paper references.

\subsection{Future work}
By providing a systematic framework and set of concrete (online) tools to allow for a structured understanding of generalisation, we believe we have set the necessary first step towards making state-of-the-art generalisation testing the new status quo in NLP.
We hope that our taxonomy will prove useful in clarifying what type of generalisation is useful in which scenario; that it will help researchers define and characterise generalisation studies, systematically registering them with our proposed evaluation cards; and that our online overview of generalisation studies will continue to provide a comprehensive picture of what happens in the field of generalisation.
Still, our work is by no means intended to be the end of the road.
For example, while our taxonomy can make future generalisation research in NLP more comparable, structured and carefully designed, and while our survey suggests interesting research directions, this work does not provide standardised data or procedures for generalisation testing.
We envision that important generalisation tests should be hosted on a shared platform, along with a leaderboard to make generalisation testing more accessible and transparent, 
and that the platform should not be controlled by a single group of people but by a larger community of NLP researchers and domain experts.
Lastly, in the same way as our thoughts on how generalisation should be evaluated have evolved with models in the past, so should such a platform continue to evolve in the future.
What we consider important to evaluate now might change next year, and when models get better at setups considered difficult today, we might discover new types of generalisation that we had not thought of before.
How we evaluate models should be reflective of this constant evolution, and which tests are prioritised should thus change along with new models and knowledge.

\section{Limitations}

Designing a coherent, consistent, and at the same time, usable taxonomy of generalisation research in NLP is a non-trivial task, which required substantial discussion among the authors.
We finish this paper by reporting the main decisional trade-offs of our work, concerning the definition of the taxonomy, the annotation process and the selection of papers to review.

First, we designed this taxonomy and its set of axes to highlight theoretically important but also practically functional distinctions between generalisation studies.
In doing so, we opted for relatively coarse axes values, which allow drawing higher-level conclusions about the field as a whole.
At the same time, this sometimes groups together papers that could be regarded separately.
An already discussed example are the studies with a pretrain-train locus, which by definition all share that they include more than one training stage and investigate generalisation in the first one.
This category thus contains both papers that use a general-purpose pretraining objective and then finetune on different tasks and studies whose finetuning objective matches the pretraining objectives (e.g.\ studies that consider domain-adaptation in a continual learning setup).
While those differences are, at least in part, reflected on other axes, in some cases it might be helpful to distinguish those two cases more explicitly.
Something similar occurs on the shift type axis.
The three formal shift types that we consider are statistically well-grounded but shifts of the same type can still largely vary.
Whether the distance between two distributions is small or large might make a substantial difference in the difficulty of the generalisation problem, which is something that is currently not reflected in our taxonomy.
Although quantifying differences between distributions is often problematic in practice, we believe that adjusting the taxonomy to capture the difficulty of generalising to a particular shift could be helpful in the future.
%
More generally, we imagine that future experimental paradigms might call for the addition of values on some of the axes, or even the addition of new axes. 
We are already observing, for example, that new studies include increasingly often more than three modelling distributions. Our taxonomy can be naturally extended to account for modelling pipelines with an arbitrary number of learning stages.

A second limitation concerns the labelling and characterisation of individual studies.
In the description of the axes and their different values, we aimed to be as comprehensive and precise as possible.
In practice, however, there are always cases in which the actual category of a paper is debatable.
Sometimes this occurs because the paper itself is not clear about what exactly it attempts to evaluate or about its motivation; we hope that our taxonomy will reduce the number of such cases in the future.
In other instances, it is simply difficult to apply some concepts or distinctions, despite their theoretical sharpness, to concrete studies.
A clear example of this challenge is the shift type.
In theory, $p(x)$, $p(y|x)$ and $p(y)$ are clearly defined concepts; in practice, it is usually impossible to estimate the actual difference between two (natural) distributions.
Some might even argue that, in practice, train and test sets are virtually always distributionally different.
For the purpose of systematising generalisation testing and characterising experiments, however, this is not a useful observation.
In our taxonomy design and annotations, we aimed to make distinctions that we deemed useful, rather than relying on ``true'' but unknown differences between distributions.

Thirdly, in our paper selection and annotation, we deliberately excluded a few types of papers.
For example, we did not include any studies that considered more than one modality.
While we believe they are interesting to consider from a generalisation perspective, they are also more difficult to characterise within a single taxonomy, as they involve more distributions (with sometimes very different support) and thus more distribution shifts.
We consider including such papers a compelling step for future work.
Another set of papers that we excluded are those that do not conduct behavioural experiments but look at the generalisability of representations (e.g.\ probing papers).
We do not see any a priori reason that they could not be characterised by our taxonomy, and we believe this would be a valuable enterprise.
In particular, although marking the difference between behavioural and representational experiments might require updating the taxonomy, a comparison of behavioural and representational experiments with the same axis values might make for an interesting meta-study.

A last critical observation that we would like to make is that our work builds on the assumption that strong generalisation skills are considered crucial for models of NLP.
While we generally believe this to be true, there might be cases where generalisation is not in fact needed.
One could provocatively argue that for LLMs trained on extremely large English data sets, practically speaking the vast majority of application scenarios is close to i.i.d.\ and that complex forms of generalisation are thus not needed.
We abstain from judging whether and when this holds but argue that when researchers believe that their setup requires no generalisation, they should clearly state so and explain why that is the case.

\section*{Acknowledgements}

We thank Adina Williams, Armand Joulin, Elia Bruni, Lucas Weber, Robert Kirk and Sebastian Riedel for providing us feedback on various stages of this draft, and Gary Marcus for providing detailed feedback on the final draft of this paper.
We thank Elte Hupkes for making the app that allows searching through references, and we thank Daniel Haziza and Ece Takmaz for other contributions to the website.

\bibliography{anthology,lit_review,darpa}
\bibliographystyle{acl_natbib}

\onecolumn

\appendix
\section{Annotation setup}\label{appendix:setup}

In this section, we describe the procedures we used for the selection of the papers in our review and their annotation.
\subsection{Paper selection}
An initial selection of manuscripts was made through a substantive preliminary literature review by the main authors of this paper.
We then carried out a search through the ACL anthology.
We started by retrieving all papers that have the (sub)words \emph{generalisation}, \emph{generalization}, \emph{generalise} or \emph{generalize} in their title or abstract.
In \cref{fig:generalisation_over_time}, we see that the number of papers with those keywords grew substantially over time, both in absolute and relative terms.
We manually checked the abstracts and titles of the resulting papers to remove those that were not, in fact, addressing a generalisation question (for instance, because they proposed a generalisation of a \emph{method}, or because they used random train--test splits).
Furthermore, we restricted ourselves to papers with one modality.
We then annotated the resulting papers using the taxonomy presented in the previous sections.
During the annotation process, we sometimes removed entries that upon further reading did not contain generalisation experiments, and we duplicated entries that contained multiple experiments with different values on one of our axes.
The findings presented in this section encompass in total \Nentries generalisation experiments, presented in \Npapers papers.
The full list of papers can be found in the second bibliography at the end of this paper, as well as on our website\footnote{\url{https://genbench.github.io/references}}.
While the conclusions in this -- static -- paper pertain only to this specific selection of papers, we intend to keep expanding the number of entries on our website with existing papers we missed or as new generalisation papers are published.

\subsection{Annotation}
\begin{figure}
    \centering
    \includegraphics[width=\textwidth]{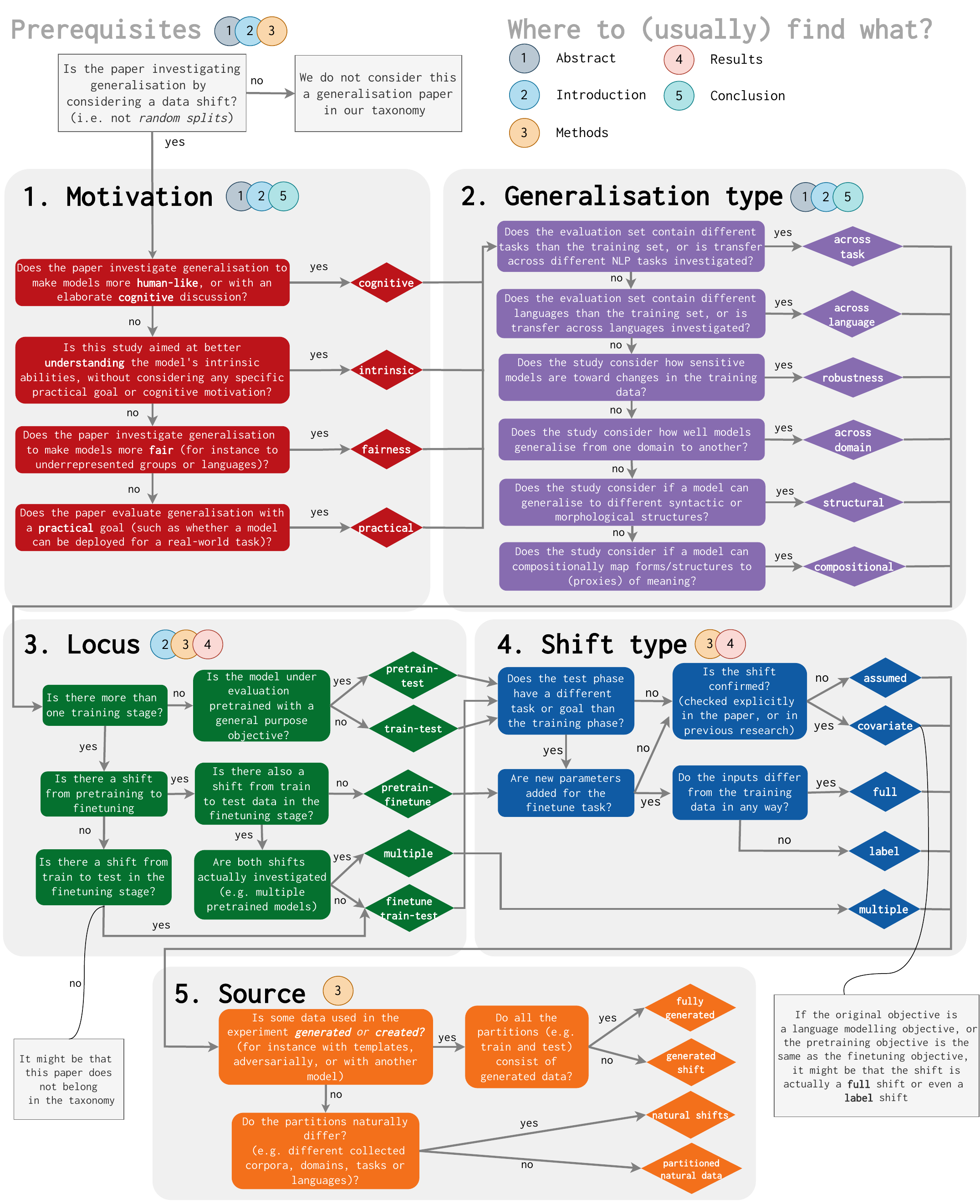}
    \caption{A graphical representation of our annotation process and an indication of where in a paper you might find the information required to complete the annotation. One paper can potentially contain multiple generalisation questions -- e.g.\ both cross-domain and cross-task generalisation, or both generated shifts and splits using natural data. In that case, the diagram has to be walked through twice. Of course, the diagram is an aid that helps characterise papers but also simplifies the full taxonomy. On our website, we keep track of common questions that arise when using the diagram to characterise papers in an FAQ.}\label{fig:annotation-diagram}
\end{figure}

The annotation of all selected papers was done collectively by the authors of this article.
Each paper was given five labels by a first annotator, one for every axis of our taxonomy, and these labels were then checked by a second annotator.
Disagreements were discussed among the two annotators, and for unresolved cases, a third annotator was used.
As a guide, we used the diagram presented in \cref{fig:annotation-diagram}.
An FAQ with common questions that occurred while using this diagram, which intends to capture our taxonomy but is naturally a simplified version of it, can be found on our website.
In addition to the taxonomy axes values, we also annotated which task(s) the studies considered.
If a paper performed the same experiment with multiple different tasks, we label it \emph{multiple tasks}, use the overarching category (e.g.\ \emph{NLU}) when possible, or mark it as \emph{multitask} if the purpose is to show that a paper can do those all at the same time.
If a paper contained multiple studies with different values on the same axis -- e.g.\ a paper considers both cross-domain and compositional generalisation or uses both natural shifts and synthetic data -- we record those experiments separately.

\newpage
\section{Evaluation cards}\label{sec:eval_cards}

In the main text of this paper, we have argued several times that we believe standardisation is an important requirement for state-of-the-art evaluation in NLP.
To further push in this direction, we propose, on top of our taxonomy, a standardised way to indicate what kind of generalisation experiments a paper reports: \emph{evaluation cards}.
Evaluation cards (for an example, see \cref{fig:eval_card}) allow visualising all generalisation experiments conducted in a study in a comprehensive way and thus to easily shows how extensively a model was evaluated.
In contrast to our review, in which multiple loci and shifts are grouped under one category for visualisation and analysis purposes, the evaluation cards do allow recording which shifts and loci are investigated in the same experiments.
In the example card in \cref{fig:eval_card}, for instance, the experiment indicated with the triangle $\bigtriangleup$ considers a covariate shift in the finetune stage (from one language to another), but through investigating multiple pretrained models it also investigates a label shift from pretraining to training.
On our \href{https://genbench.org/eval_cards/}{website}, we provide a tool to generate (one or two-column) evaluation cards, that authors can include in the appendix of their papers.
In \cref{fig:eval_card_interface}, we show how this tool is rendered in our web interface.

\begin{figure}
    \centering
    \includegraphics[width=0.75\textwidth]{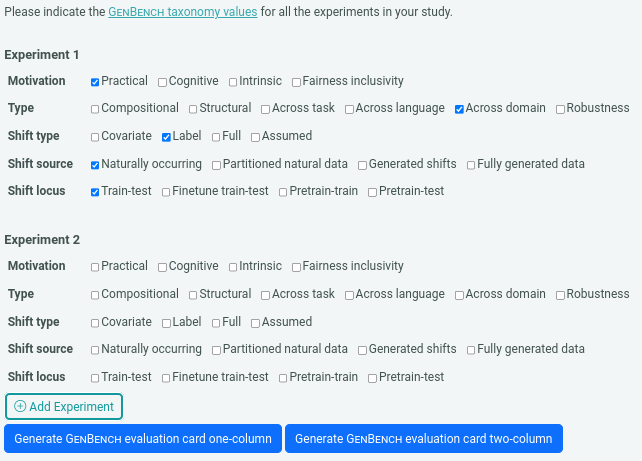}
    \caption{Our online interface generates LaTeX code for one- or two-column evaluation cards.\looseness-1}\label{fig:eval_card_interface}
\end{figure}

\newpage
\section{Author contributions}
\label{ap:author_contributions}
To recognise individual author contributions, we detail those contributions below, following the \textit{Contributor Roles Taxonomy} (CRediT)\footnote{\url{https://www.elsevier.com/authors/policies-and-guidelines/credit-author-statement}} introduced by Elsevier.
Authors are listed in the order they appear on the author list.\\
\footnotetext[21]{Work done outside of Amazon}

\begin{table}[h!]
\renewcommand\arraystretch{1.2}
\centering
\begin{tabular}{|ll|}
    \hline
    Dieuwke Hupkes & \textit{\specialcell{Conceptualisation, Methodology, Software, Validation,\\
                                  Formal Analysis, Investigation, Data Curation,\\
                                  Writing -- Original Draft, Writing -- Review \& Editing,\\
                                  Visualisation, Supervision, Project Administration}} \\
    \hline
    Mario Giulianelli & \textit{\specialcell{Conceptualisation, Methodology, Validation, Formal Analysis,\\
                                     Investigation, Data Curation, Writing -- Original Draft,\\
                                     Writing -- Review \& Editing, Project Administration}}  \\
    \hline
    Verna Dankers & \textit{\specialcell{Conceptualisation, Software, Formal Analysis, Data Curation,\\
                                 Writing -- Review \& Editing, Visualisation}} \\
    \hline
    Mikel Artetxe & \textit{\specialcell{Conceptualisation, Data Curation, Validation, \\
                                 Writing -- Original Draft, Writing -- Review \& Editing}} \\
    \hline
    Yanai Elazar & \textit{\specialcell{Conceptualisation, Data Curation, Writing -- Original Draft\\
                                Writing -- Review \& Editing}} \\
    \hline
    Tiago Pimentel & \textit{\specialcell{Conceptualisation, Data Curation,\\ Writing -- Original Draft,
                                  Writing -- Review \& Editing} }\\
    \hline
    Christos Christodoulopoulos\footnotemark & \textit{\specialcell{Conceptualisation, Data Curation, Writing -- Original Draft}} \\
    \hline
    Karim Lasri & \textit{\specialcell{Conceptualisation, Data Curation, Writing -- Original Draft,\\ Visualisation}} \\
    \hline
    Naomi Saphra & \textit{\specialcell{Conceptualisation, Writing -- Original Draft}} \\
    \hline
    Arabella Sinclair & \textit{\specialcell{Conceptualisation, Data Curation}} \\
    \hline
    Dennis Ulmer & \textit{\specialcell{Data Curation, Writing -- Review \& Editing, Visualisation}} \\
    \hline
    Florian Schottmann & \textit{\specialcell{Data Curation}} \\
    \hline
    Khuyagbaatar Batsuren & \textit{\specialcell{Data Curation}} \\
    \hline
    Kaiser Sun & \textit{\specialcell{Data Curation}} \\
    \hline
    Koustuv Sinha & \textit{\specialcell{Data Curation}} \\
    \hline
    Leila Khalatbari & \textit{\specialcell{Data Curation}} \\
    \hline
    Maria Ryskina & \textit{\specialcell{Software, Data Curation}} \\
    \hline
    Rita Frieske & \textit{\specialcell{Data Curation}} \\
    \hline
    Ryan Cotterell & \textit{\specialcell{Data Curation, Writing -- Review \& Editing}} \\
    \hline
    Zhijing Jin & \textit{\specialcell{Data Curation}} \\
    \hline
\end{tabular}
    \caption{Individual author contributions using CRediT.}
\end{table}

\newpage
\section{Multi-lingual benchmarks}\label{appendix:multilingual_benchmarks}

While the field of multilingual modelling is vast and associated with many interesting generalisation questions, papers in this area do not often focus explicitly on generalisation.
In this section, we provide a list of the most important available multilingual benchmarks which can be used to evaluate cross-lingual generalisation.
Multilingual benchmarks or datasets are created in a variety of ways.
Several benchmarks are created by translating monolingual benchmarks into different languages, usually through a professional translation service \citep{ponti-etal-2020-xcopa,li2021xglm,mostafazadeh-etal-2016-corpus,yang-etal-2019-paws,zhang-etal-2019-paws,conneau-etal-2018-xnli,williams-etal-2018-broad,ebrahimi-etal-2022-americasnli,artetxe-etal-2020-cross,lewis-etal-2020-mlqa,longpre-etal-2021-mkqa,xu-etal-2020-end,li-etal-2021-mtop,fitzgerald2022massive}.
Other multilingual benchmarks, instead, have been built by separately annotating each language via its native speakers \citep[e.g.][]{adelani-etal-2021-masakhaner, clark-etal-2020-tydi,asai-etal-2021-xor,muller2021genqa}.
Yet another way to construct multilingual benchmarks is to leverage existing resources that cover multiple languages.
For instance, Wikipedia has been used as a resource to derive multilingual benchmarks \citep{pan-etal-2017-cross,rahimi-etal-2019-massively,botha-etal-2020-entity,liu-etal-2019-xqa}, and several multilingual summarisation datasets have been created by extracting article-summary pairs from online newspapers or how-to guides \citep[e.g.][]{nguyen-daume-iii-2019-global,scialom-etal-2020-mlsum,ladhak-etal-2020-wikilingua,hasan-etal-2021-xl,varab-schluter-2021-massivesumm}.
Various linguistic resources have also been exploited:
for instance, the Universal Dependencies treebank \citep{nivre-etal-2020-universal} has been used to evaluate cross-lingual part-of-speech tagging, and multilingual WordNet and Wiktionary have been used to build XL-WiC \citep{raganato-etal-2020-xl}, an extension of WiC \citep{pilehvar-camacho-collados-2019-wic} that reformulates word sense disambiguation in 12 languages as a binary classification task.
Finally, in the same spirit of GLUE and SuperGLUE for English, several aggregated benchmarks include selected sets of benchmarks previously proposed by others \citep[e.g.][]{hu2020xtreme, ruder-etal-2021-xtreme,liang-etal-2020-xglue,wang2022benchmarking}, which allow for evaluating cross-task and cross-language generalisation simultaneously.

\begin{figure}[h!]
    \centering
    \includegraphics[width=1.0\textwidth]{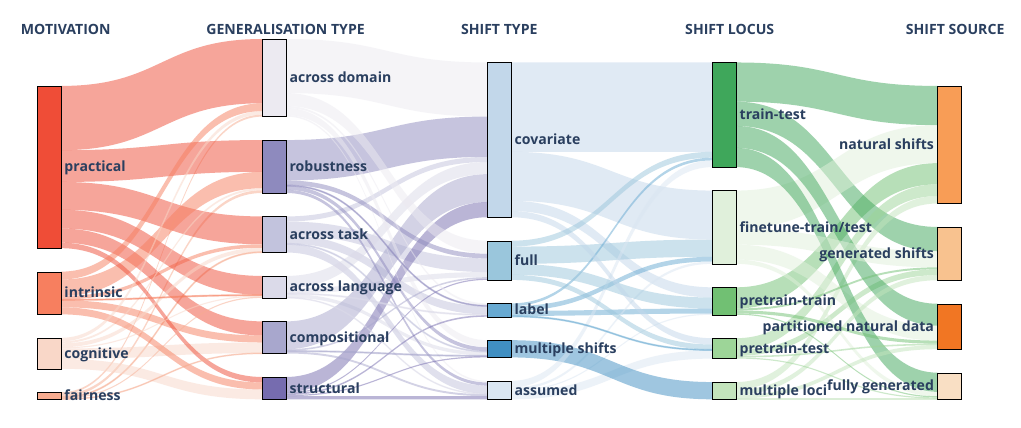}
    \caption{An overview figure of our literature review, including interactions. An interactive version of this plot can be found on our website \url{https://genbench.github.io/visualisations}. For more detailed explanations and analyses, we refer to \cref{sec:survey}.}\label{fig:sankey-overview}
\end{figure}

\newpage
\section{A concise summary of our taxonomy}\label{appendix:taxonomy_summary}

For the convenience of the reader, in this section of the supplementary materials we provide a concise summary of our generalisation taxonomy.
The taxonomy we propose is based on a detailed analysis of a large number of existing studies on generalisation in NLP, and it includes the main five axes along which those studies differ.
The five axes capture different aspects of generalisation studies, that together form a comprehensive picture of the motivation and goal of the study and provide information on important choices in the experimental setup.

The first axis of our generalisation taxonomy (\cref{sec:motivations}) is the high-level \textbf{motivation} for the study.
The motivation of a study impacts or even determines what type of generalisation is desirable, as well as what kind of conclusions can be drawn from a model's display or lack of generalisation.
Furthermore, the motivation of a study shapes its experimental design.
It is therefore important for researchers to be explicitly aware of it, to ensure that the experimental setup aligns with the questions they seek to answer.
We consider four different types of motivations: the \emph{practical} motivation, the \emph{cognitive} motivation, the \emph{intrinsic} motivation, and the \emph{fairness and inclusivity} motivation.

The second axis in our taxonomy (\cref{sec:generalisation_target}) indicates the \textbf{type of generalisation} the test is addressing.
This axis describes on a high level what exactly it is that a generalisation test is intended to capture, rather than considering why or how, making it one of the most important axes of our taxonomy.
In the literature, we have found six main types of generalisation: \emph{compositional} generalisation, \emph{structural} generalisation, \emph{cross-task} generalisation, \emph{cross-lingual} generalisation, \emph{cross-domain} generalisation, and \emph{robustness} generalisation.

The third axis in our taxonomy (\cref{sec:data_shifts}) describes what kind of \textbf{data shift} is considered in the generalisation test.
This axis adds a statistical interpretation to our taxonomy and derives its importance from the fact that data shift plays an essential formal role in defining and understanding generalisation from a statistical perspective, as well as from the fact that different types of shifts are best addressed with different kinds of experimental setups.
On the data shift axis, we consider three shifts which are well-attested in the literature: \emph{covariate shift}, \emph{label shift} and \emph{full shift}.
We further include two additional types of shift -- \textit{assumed shift} and \textit{multiple shifts} -- to account for studies that cannot be labelled with any of the three main shift types.

In the fourth axis of our taxonomy (\cref{sec:shift_sources}), we consider what is the \textbf{source} of the data shift used in the experiment.
The source of the data shift determines how much control the experimenter has over the training and testing data and, consequently, what kind of conclusions can be drawn from an experiment.
We distinguish four different sources of shifts: \emph{naturally occurring shifts}, \emph{artificially partitioned natural corpora}, \emph{generated shifts} and \emph{fully generated datasets}.

In the last axis of our taxonomy (\cref{sec:shift_locus}), we consider what is the \textbf{locus} of the data shift, or, in other words, for what part of the modelling pipeline generalisation is investigated.
The locus of the shift, together with the shift type, forms the last piece of the puzzle, as it determines what part of the modelling pipeline is investigated and thus the kind of generalisation question that can be asked.
On this axis, we consider shifts between all stages in the contemporary modelling pipeline -- pretraining, training and testing -- as well as studies that consider shifts between multiple stages simultaneously.

\newpage

\begingroup
\def\refname{}
\vspace{-28mm}
\includepdf[pages={2-}, pagecommand={\thispagestyle{plain}\hypertarget{appendix:references}}, addtotoc={2, section, 1, List of publications included in our review, pool1}]{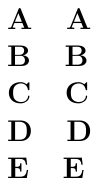}
\endgroup

\label{sec:appendix}

\end{document}